\documentclass[a4paper,fleqn]{cas-sc}

\usepackage[authoryear]{natbib}
\usepackage{booktabs}
\usepackage{graphicx}
\usepackage{float}
\usepackage{cas-common}
\usepackage{hyperref}
\usepackage{soul}
\usepackage{soulpos}
\sethlcolor{yellow}
\usepackage[most]{tcolorbox}
\newtcolorbox{highlighted}{
  colback=yellow!20,        
  boxrule=0pt,              
  breakable,                
  left=2pt, right=2pt, top=2pt, bottom=2pt
}

\def\tsc#1{\csdef{#1}{\textsc{\lowercase{#1}}\xspace}}
\tsc{WGM}
\tsc{QE}

\begin{document}
\let\WriteBookmarks\relax
\def\floatpagepagefraction{1}
\def\textpagefraction{.001}

\shorttitle{SpaRED}    

\shortauthors{D. Ruiz}  

\title [mode = title]{Completing Spatial Transcriptomics Data for Gene Expression Prediction Benchmarking}



%

\cortext[1]{Corresponding author}



\affiliation[1]{organization={Center for Research and Formation in Artificial Intelligence, Universidad de los Andes, Colombia},
            addressline={Carrera 1 No. 18a-12}, 
            city={Bogotá},
            postcode={111711}, 
            country={Colombia}}

\author[1]{Daniela Ruiz}[orcid=0000-0001-6636-1173]
\ead{da.ruizl1@uniandes.edu.co}
\credit{Project administration, Data Curation, Methodology, Formal analysis, Research, Software, Visualization, Writing}

\author[1]{Paula Cárdenas}[orcid=0009-0005-1185-548X]
\ead{p.cardenasg@uniandes.edu.co}
\credit{Data Curation, Methodology, Formal analysis, Research, Software, Writing}

\author[1]{Leonardo Manrique}[orcid=0009-0008-9428-6009]
\ead{dl.manrique@uniandes.edu.co}
\credit{Data Curation, Formal analysis, Research, Software, Visualization, Writing}

\author[1]{Daniela Vega}[orcid=0009-0002-9731-7591]
\ead{d.vegaa@uniandes.edu.co}
\credit{Data Curation, Formal analysis, Research, Software, Visualization, Writing}

\author[1]{Gabriel M. Mejia}[orcid=0000-0003-4382-6390]
\ead{gm.mejia@uniandes.edu.co}
\credit{Conceptualization, Data Curation, Methodology, Formal analysis, Research, Software}

\author[1]{Pablo Arbeláez}[orcid=0000-0001-5244-2407]
\ead{pa.arbelaez@uniandes.edu.co}
\credit{Conceptualization, Funding acquisition, Supervision, Writing}

 
\begin{abstract}
Spatial Transcriptomics is a groundbreaking technology that integrates histology images with spatially resolved gene expression profiles. Among the various Spatial Transcriptomics techniques available, Visium has emerged as the most widely adopted. However, its accessibility is limited by high costs, the need for specialized expertise, and slow clinical integration. Additionally, gene capture inefficiencies lead to significant dropout, corrupting acquired data. To address these challenges, the deep learning community has explored the gene expression prediction task directly from histology images. Yet, inconsistencies in datasets, preprocessing, and training protocols hinder fair comparisons between models. To bridge this gap, we introduce SpaRED, a systematically curated database comprising 26 public datasets, providing a standardized resource for model evaluation. We further propose SpaCKLE, a state-of-the-art transformer-based gene expression completion model that reduces mean squared error by over 82.5\% compared to existing approaches. Finally, we establish the SpaRED benchmark, evaluating eight state-of-the-art prediction models on both raw and SpaCKLE-completed data, demonstrating SpaCKLE substantially improves the results across all the gene expression prediction models. Altogether, our contributions constitute the most comprehensive benchmark of gene expression prediction from histology images to date and a stepping stone for future research on Spatial Transcriptomics.
\end{abstract}

\begin{graphicalabstract}
\includegraphics[width=0.99\textwidth]{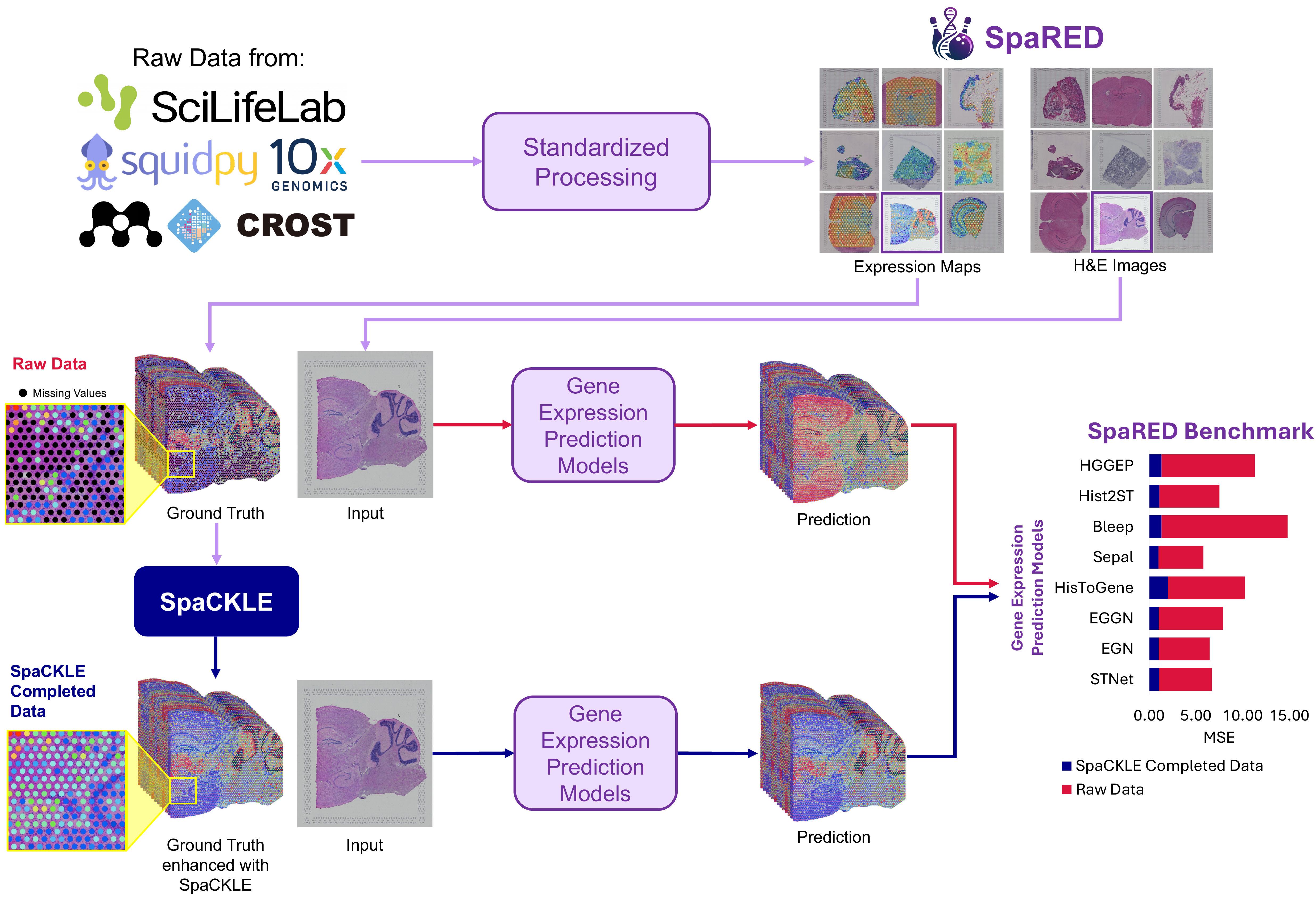}
\end{graphicalabstract}

\begin{highlights}
\item Release of SpaRED, a preprocessed set of 26 Spatial Transcriptomics datasets.
\item Propose SpaCKLE, a transformer-based method to complete missing gene expressions.
\item Benchmarking of eight common models for gene expression from histology images.
\item Open-source database and gene completion model with an easy-to-use Python library.

\end{highlights}



\maketitle

\section{Introduction}\label{Introduction}

Spatial Transcriptomics (ST) is an emerging technology that precisely localizes gene expression profiles within histological images \citep{jiang2023generalization}. While histology analysis is the gold standard for the diagnosis of many diseases \citep{xie2023spatially}, transcriptomics unlocks molecular insights that unveil causal pathways behind pathologies \citep{zeng2022spatial,jiang2023generalization}. Beyond disease research, ST has broad applications in developmental biology, enabling the study of tissue formation, cellular differentiation, and organogenesis with spatial resolution \citep{choe2023advances}. Additionally, ST is valuable in regenerative medicine and tissue engineering, guiding the design of biomaterials and cell-based therapies through a deeper understanding of gene expression patterns in healthy and regenerating tissue \citep{lammi2024spatial}. By integrating histology with transcriptomics, ST opens a new spectrum of possibilities to understand tissue structure and mechanistic insights into various biological processes \citep{wang2023crost}. 

\begin{figure*}[t]
    \includegraphics[width=0.99\textwidth]{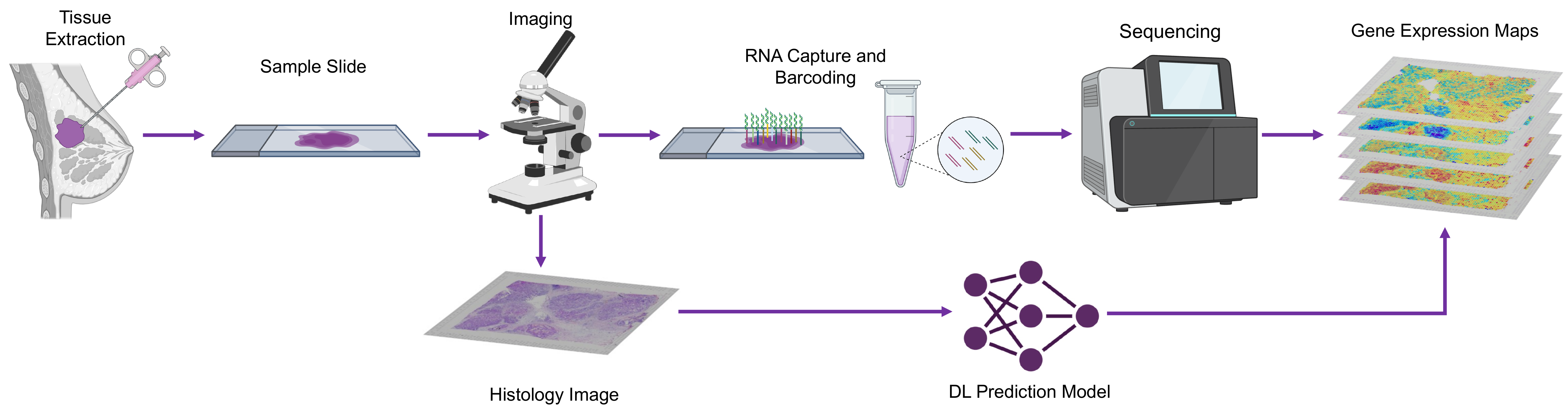}
    \caption{\textbf{Spatial Transcriptomics Overview.} The process begins with placing a fresh frozen tissue on a slide for imaging. The tissue is then fixed and permeabilized, releasing RNA, which binds to capture probes for gene expression profiling. Barcoded cDNA is synthesized from the captured RNA, generating sequencing libraries. The sequencing data is then processed using specialized analysis software to create spatially resolved gene expression maps. However, these experiments are costly and require specialized equipment. Deep learning (DL) models for gene expression prediction offer an alternative by generating gene expression maps solely from histology images, streamlining the process and making Spatial Transcriptomics more accessible.}
    \label{fig:ST_overview}
\end{figure*}

As with any emerging technology, multiple variations of ST are currently available and under continuous development \citep{slide_seq_v2,merfish,10x_visium}. Notably, as demonstrated by the number of entries in the comprehensive ST repository \citep{wang2023crost}, Visium \citep{10x_visium} has emerged as the most widely used ST technology. The workflow of this technology is depicted in Fig. \ref{fig:ST_overview} and begins with the preparation of the tissue, where the sample is embedded, sectioned, and placed on a slide with designated capture areas. Next, staining and imaging are performed using standard histological techniques to visualize tissue structures. Once imaged, the tissue is permeabilized, allowing mRNA to be released. Then, this mRNA is captured using barcoded oligonucleotides, enabling spatial mapping of gene expression. A reverse transcription reaction is then used to synthesize cDNA from the captured mRNA, which is subsequently processed into sequencing libraries. Finally, specialized analysis software processes the sequencing data, generating spatially resolved gene expression maps for visualization and interpretation \citep{10x_visium}.

Despite its advantages, this approach presents key challenges: high costs, the need for domain expertise, and slow adoption in clinical settings, limiting its accessibility in routine diagnostics \citep{pang2021leveraging}. In addition to these challenges, on the technical side, it inherits data capturing issues from bulk and single-cell transcriptomics \citep{pham2023robust,avsar2023comparative}. This problem is known as dropout and corresponds to the failure to detect transcripts even though they are present in the source tissue. In practice, this phenomenon appears as pepper noise in gene expression maps, often requiring single-cell reference datasets to compensate for missing data \citep{avsar2023comparative}.

Acknowledging these challenges, the deep learning community has delved into democratizing ST by studying gene expression prediction from histology images \citep{jiang2023generalization}. By bypassing the need for specialized sequencing, these approaches offer a more accessible and scalable alternative, enabling subjects to obtain molecular insights of a tissue from a standard biopsy image. Leveraging the abundance of public Visium data, multiple deep learning models have emerged to tackle this task \citep{he2020integrating,pang2021leveraging,yang2023exemplar,yang2024spatial,xie2023spatially,zeng2022spatial,mejia2023SEPAL}. Although these methods consistently report favorable results against the latest state of the art, differences in datasets, preprocessing strategies, and training hyperparameters hinder fair comparisons and compromise the validity of new findings.

In our previous MICCAI paper titled "Enhancing Gene Expression Prediction from Histology Images with Spatial Transcriptomics Completion" \citep{mejia2024enhancing}, we introduced initial efforts to address the limitations discussed above. In this work, we substantially build upon and refine those initial contributions. First, we enhance the methodology by introducing comprehensive ablation studies to support our design choices for SpaCKLE, including the contribution of data pre-completion, the integration of visual features, the effect of incorporating context genes information, and the impact of neighborhood size. Second, we broaden the SpaRED benchmark by adding the state-of-the-art model HGGEP \citep{li2024gene} and systematically evaluating its performance across all 26 datasets. Third, we provide a more comprehensive analysis with additional qualitative and statistical results for both our completion model and the SpaRED Benchmark, offering more profound insights into SpaCKLE's performance and a more detailed comparative evaluation of existing gene expression prediction models.

Our key contributions can be summarized as follows.

\begin{enumerate}
    \item We systematically compile, curate, and standardize 26 public ST datasets into the \textbf{Spa}tially \textbf{R}esolved \textbf{E}xpression \textbf{D}atabase (\textbf{SpaRED}), an extensive Visium resource encompassing human and mouse samples from nine tissue types.
    \item To address the dropout problem, we introduce \textbf{Spa}tial transcriptomics \textbf{C}ompletion with \textbf{K}nowledge from the \textbf{L}ocal \textbf{E}nvironment (\textbf{SpaCKLE}), a transformer-based model inspired by the unrivaled power of self-attention mechanisms for next token prediction in natural language processing \citep{dosovitskiy2020image}. Notably, SpaCKLE surpasses existing gene completion approaches, achieving a relative 82.5\% MSE reduction compared to the median method. 
    \item  We establish the \textbf{SpaRED benchmark}, evaluating \textbf{eight} state-of-the-art prediction models on both raw and SpaCKLE-completed data. This benchmark exposes the proximity in performance across all the models we study and the need for exploring new approaches in this task. Moreover, our benchmark also demonstrates that SpaCKLE significantly enhances gene expression prediction performance across all tested models.
\end{enumerate}

To ensure the reproducibility of our experiments and facilitate the implementation of SpaCKLE, we provide the SpaRED library, available at \href{https://spared.readthedocs.io/en/main/api.html}{PyPI}. Additionally, we present a \href{https://bcv-uniandes.github.io/spared\_webpage/}{web platform} to explore SpaRED data, access key statistics, and download both raw and processed datasets.

\section{Related Work}\label{Related Work}

\subsection{Integrated Databases}\label{Integrated Databases}

Recent advances in ST have led to the development of multiple databases. For instance, CROST \citep{wang2023crost} is a comprehensive repository with 1033 ST samples from 8 species, 35 tissues, and 56 diseases. Other databases include SpatialDB, Aquila, SPASCER, SODB, and STomicsDB \citep{wang2023crost}, each offering unique datasets and analytical tools. Although these databases facilitate advanced spatial analyses, they are not specifically designed for the expression profile prediction task.

With the increasing publication of individual datasets obtained with ST technology, database integration has become increasingly important. Similar to the data integration efforts that we present in this work, HEST-1k \citep{hest1k} is a recent database that compiles over 1,229 spatial transcriptomic profiles paired with H\&E-stained whole-slide images (WSIs). Like SpaRED, HEST-1k provides essential metadata such as organ type and species, and it also extends this information by including disease status, treatment information for cancer samples, and details on the ST technology used. These additional metadata fields facilitate structured analyses across diverse biological conditions. While HEST-1k also presents a python library that provides the possibility of performing data processing, they report not including batch effect correction during their inherent data preprocessing, which may become a limitation for the downstream use of the ST data. Similarly,  the newly introduced STimage-1K4M \citep{chen2024stimage} database contains a diverse collection of 1,149 ST slides, encompassing 4,293,195 spots. Alongside this data, it includes pathologist annotations for 71 slides, providing a valuable resource for assessing clustering methods and dimensionality reduction techniques. Nonetheless, STimage-1K4M lacks any data processing, potentially hindering its use in certain applications.

SpaRED tackles the limitations mentioned above by implementing best practices in bioinformatics analysis, including the selection of Moran genes, standardization of reference genomes, TPM normalization, and batch effect correction. These steps enhance data reliability and comparability, making SpaRED particularly useful for clinical applications. Furthermore, SpaRED optimally organizes the association between WSIs and gene transcripts, ensuring a structured framework for tasks that rely on multimodal relationships, such as predicting gene expression from histology images. 

\subsection{Completion strategies}\label{Completion strategies}

Several strategies have been proposed to address missing data in spatial transcriptomics, broadly falling into two categories: reference-based and reference-free methods.

Reference-based methods integrate spatial transcriptomics with matching single-cell RNA-seq datasets to infer missing gene expression and enhance resolution. For instance, Tangram \citep{biancalani2021deep} aligns spatial and single-cell transcriptomic profiles to map gene expression from the single-cell domain onto tissue. SpaGE \citep{abdelaal2020spage} projects spatial data into a latent space constructed from single-cell references to predict unmeasured genes. Seurat \citep{stuart2019comprehensive}, Harmony \citep{korsunsky2019fast}, LIGER \citep{welch2019single}, gimVI \citep{lopez2019joint}, and stPlus \citep{shengquan2021stplus} follow similar principles, combining multimodal alignment or probabilistic modeling to impute spatial gene expression using external scRNA-seq data.

While effective, these methods require carefully curated, high-quality single-cell datasets from the same tissue and condition—resources that are often difficult to obtain. Moreover, utilizing an scRNA-seq reference, increases the resources needed to clean the ST dataset at the risk of inducing batch effects due to dissimilar sequencing technologies \citep{marel2024navigating}. Additionally, reference-based models depend on alignment quality, which remains suboptimal despite ongoing advancements, potentially introducing bias in the completed data \citep{yan2024integration}.

Reference-free approaches, in contrast, rely exclusively on the spatial structure and expression context within each ST slide. For example, SEPAL \citep{mejia2023SEPAL} uses a modified adaptive median filter to replace dropout values with the median expression in a local circular region; if the region lacks sufficient data, the method falls back to the global median. Alternatively, stLearn \citep{pham2023robust} uses genetic and morphological similarity to adjust existing spots or predict gene expression for missing values. 

SpaCKLE is a reference-free completion method, which stands out from alternatives by leveraging the complete genetic profile of adjacent spots and taking advantage of the transformer capacity to predict missing values. SpaCKLE captures local gene-gene and spot-spot dependencies, allowing it to complete missing values using only intrinsic spatial information. This makes it especially suited for Visium datasets lacking a corresponding single-cell reference, offering greater flexibility and broader applicability.

\subsection{Gene Expression Prediction Benchmarks}
The gene expression prediction task for ST corresponds to the problem of automatizing the computation of the genetic profile of the spots in a tissue to reduce the costs and limitations of traditional ST data collection. This processing involves the use of computer vision techniques that obtain a histology image and output the expressions that compose the volume of gene expression maps associated with the WSI. During the past few years, multiple Artificial Intelligence (AI) models have been proposed to tackle this task, showing the importance of presenting a benchmark that clearly and fairly highlights the differences in performance between these models.

A recent study by \cite{jiang2023generalization} reviews six deep learning methods for gene expression profile prediction, testing their performance on 3 distinct breast cancer datasets. Although the study provides a solid performance analysis, it focuses exclusively on human breast cancer tissue. Moreover, HEST-1k \citep{hest1k} also presents a benchmark for the gene expression prediction task. However, instead of evaluating models explicitly designed for gene expression prediction, they assess the effectiveness of histology foundation models in visual feature extraction. Their approach involves using embeddings from 11 different histology foundation models as input to train regression models that predict gene expression. These regression models, based on standard machine learning techniques such as XGBoost, are trained to map histology image embeddings to the log1p-normalized expression levels of preselected genes. Specifically, HEST-1k focuses on predicting the expression of the 50 most highly variable genes across only nine datasets, all derived from human cancer samples, to determine the ability of the foundation models to provide the most informative embeddings for gene expression prediction. Additionally, they evaluate model performance exclusively using the Pearson correlation coefficient (PCC), which primarily measures linear associations and may not fully capture how well predictions approximate the actual gene expression values.

Considering the benchmark strategy and focus of both \cite{jiang2023generalization} and \cite{hest1k}, we find key differences with the SpaRED benchmark we present in this work. While \cite{hest1k} includes a greater total number of datasets, slides, and spots, it reports results for only 9 datasets. In contrast, SpaRED reports results on 26 datasets, which is 2.8 times more than Jaume et al. (2024) and 8.6 times more than \cite{jiang2023generalization}. Additionally, SpaRED benchmark covers nine different tissue types from both human and mouse subjects with healthy and pathological cases, in contrast to \cite{jiang2023generalization}. Fig. \ref{fig:spared} provides detailed statistics on the number of datasets and spots for each tissue and organism, allowing for a comprehensive evaluation of model generalizability. Additionally, instead of indirectly evaluating histology foundation models through feature extraction, SpaRED directly assesses the performance of state-of-the-art models specifically designed for gene expression prediction. Finally, we include the Mean Square Error (MSE) as an additional performance metric to PCC, to provide a more detailed assessment of each method's predictive accuracy.

\section{Spatially Resolved Expression Database}\label{Spatially Resolved Expression Database}
 
\subsection{Original Datasets and Curation}\label{Original Datasets and Curation}

\begin{figure}[t]
\centering
    \includegraphics[width=0.99\textwidth]{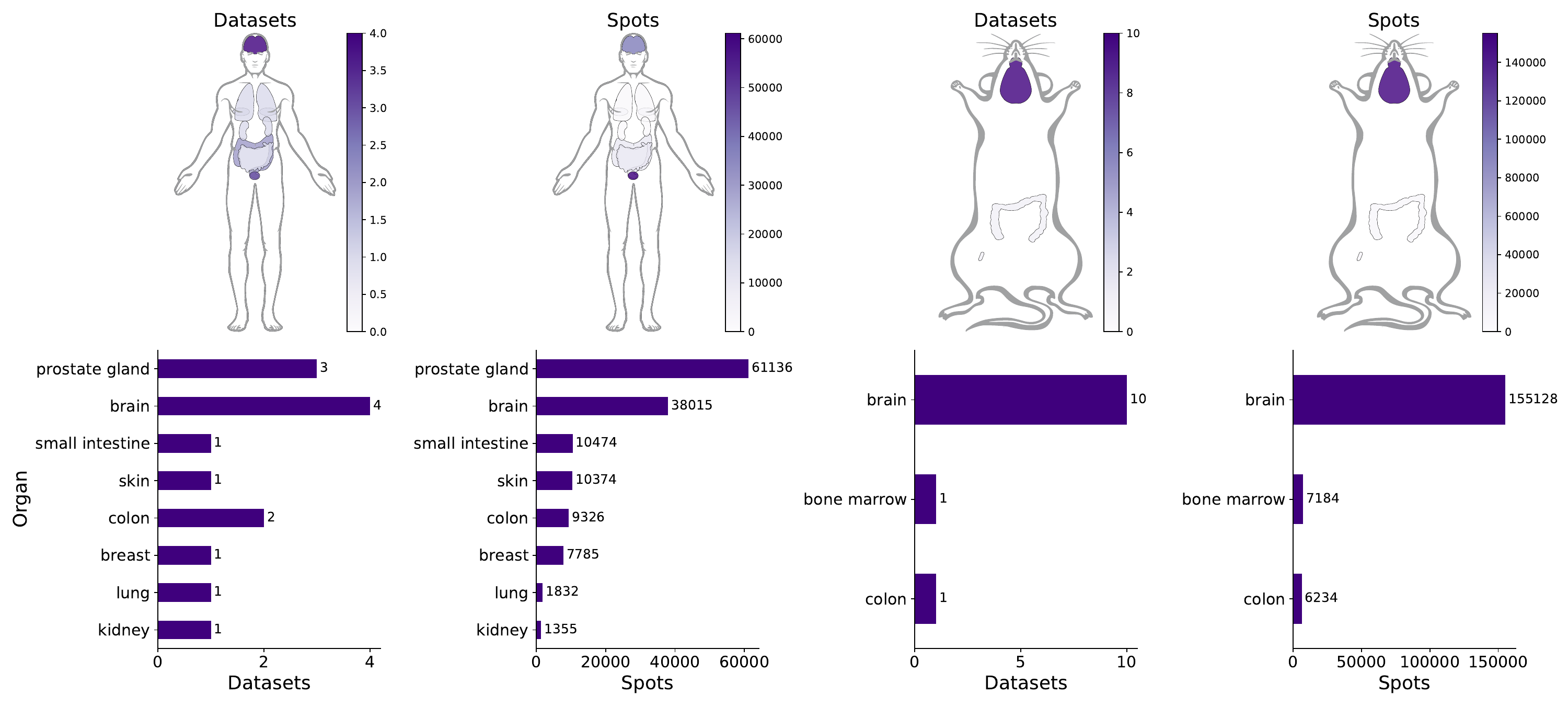}
    \caption{\textbf{SpaRED Database Statistics.} Organisms and tissues available in SpaRED, along with the number of datasets and spots available from each tissue.}
    \label{fig:spared}
\end{figure}

To build SpaRED, we collect raw data from 7 independent publications \citep{Abalo2021}, \citep{parigi2022spatial}, \citep{villacampa2021genome}, \citep{vicari2023spatial}, \citep{mirzazadeh2023spatially}, \citep{erickson2022spatially}, \citep{fan2023expansion} and complement them using 5 demonstration datasets from 10X Genomics (available through the SquidPy python package \citep{palla2022squidpy}). We only include datasets with more than one WSI and split the publications' data by tissue type, resulting in 26 distinct datasets: 14 from human and 12 from mouse, showcasing a variety of tissue samples, as illustrated in Fig. \ref{fig:spared}.  

Building on this, we define two types of generalization tasks based on the number of subjects in each dataset. Intra-subject generalization considers WSIs as consecutive sections from a single tissue and subject, whereas inter-subject generalization involves WSIs from the same tissue type but across different subjects. To ensure balanced visual distributions, we manually assign WSIs to train, validation, and test sets. In 11 out of 26 cases, a separate test set is defined. For datasets with a limited number of subjects or slides, we instead split the data into training and validation sets.  

Subsequently, we implement a structured preprocessing pipeline composed of two main stages: data filtering and data processing. In the filtering stage, we begin by setting minimum and maximum cell count thresholds \([10, 1000000]\), excluding observations that fall outside this range. We then calculate both per-slide and global expression fractions for each gene, retaining only those that meet the minimum expression threshold across spots and the entire dataset. Additionally, we filter out genes with counts outside the predefined range \([10, 1000000]\) and remove cells with zero expression across all genes. This initial filtering step is intentionally aggressive, eliminating the vast majority of low-expression or non-informative genes to improve data quality and reduce noise.

Following data filtering, we proceed with dataset processing. We normalize gene expression using Transcripts Per Million (TPM) and apply a $log_2(x+1)$ transformation. We compute Moran's I for each gene on each slide to assess spatial autocorrelation, averaging the values across all slides. Inspection of the full Moran’s I ranking revealed that genes falling at the bottom exhibited markedly higher proportions of missing data. Consequently, we retain the top 128 or 32 genes with the highest spatial autocorrelation. We select these numbers to ensure a consistently high-quality set of informative genes and computational efficiency across all datasets. However, the number of genes is customizable when using the SpaRED library for dataset creation. Finally, we apply ComBat \citep{johnson2007adjusting} batch correction to mitigate batch effects.  

As a result, the final SpaRED dataset includes 105 slides and 308,843 spots from 35 subjects. Table \ref{table:spared_details} provides a comprehensive breakdown of the dataset statistics. Moreover, it shows the proportion of missing data before and after processing. This proportion refers to the fraction of spatial spots with missing expression values for any of the 32 or 128 retained genes in each dataset. The table \ref{table:spared_details} demonstrates that our processing pipeline effectively cleans the data, substantially reducing the amount of missing values.

\begin{table}[]
\centering
\caption{Detailed overview of the SpaRED datasets, showcasing the generalization task, the number of genes analyzed, the abbreviation for each dataset, the organism studied, the tissue or disease, and the number of slides, subjects and spots. Additionally, the table presents key statistics on data corruption and missing values. The `Corrupt Spots' column shows the percentage of spots with at least one corrupted value in each dataset. `Missing data before' indicates the proportion of missing data prior to any processing, while `Missing data after' reflects the percentage remaining after processing.}

\resizebox{0.99\textwidth}{!}{
\begin{tabular}{c|c|ccccccccc}
\hline
\textbf{\begin{tabular}[c]{@{}c@{}}Generali-\\ zation\end{tabular}} & \textbf{Genes} & \textbf{\begin{tabular}[c]{@{}c@{}}Abbreviation/\\ Access\end{tabular}}  & \textbf{Organism} & \textbf{\begin{tabular}[c]{@{}c@{}}Tissue/\\ Disease\end{tabular}}  & \textbf{Slides} & \textbf{Subjects} & \textbf{Spots} & \textbf{\begin{tabular}[c]{@{}c@{}} Corrupt \\ Spots \end{tabular}} & \textbf{\begin{tabular}[c]{@{}c@{}} Missing\\ data\\before\end{tabular}} & \textbf{\begin{tabular}[c]{@{}c@{}}  Missing\\ data \\after\end{tabular}} \\ 
\hline
\multirow{7}{*}{\begin{tabular}[c]{@{}c@{}}Inter-\\ Subject\end{tabular}}  & \multirow{4}{*}{128}  &     \href{https://figshare.scilifelab.se/articles/dataset/Spatial_Multimodal_Analysis_SMA_-_Spatial_Transcriptomics/22778920}{VMB [20]}
& Mouse  & Brain  & 14 & 4 & 43804  & 100\% & 89\% & 28\%  \\
& & \href{https://data.mendeley.com/datasets/4w6krnywhn/1}{MMBR [12]} 
& Mouse & Brain  & 8  & 2  & 34583 & 100\%  & 79\%  & 20\%  \\
& & \href{https://data.mendeley.com/datasets/4w6krnywhn/1}{MHPC [12]}
& Human & Prostate cancer  & 4 & 2  & 15684 & 100\%   & 79\%  & 21\%  \\
& & \href{https://www.ncbi.nlm.nih.gov/geo/query/acc.cgi?acc=GSE169749}{PMI [15]}
& Mouse & Intestine & 2  & 2 & 6234 & 99\%  & 79\%  & 7\% \\ \cline{2-11} 
& \multirow{3}{*}{32}   &  \href{https://data.mendeley.com/datasets/xjtv62ncwr/3}{VLMB [21]} 
& Mouse & Brain  & 5  & 2  & 12202 & 100\%  & 95\%  & 29\%  \\
& & \href{https://data.mendeley.com/datasets/4w6krnywhn/1}{MHSI [12]} 
& Human & Small intestine & 4  & 2  & 10474 & 100\%  & 92\%  & 21\% \\
&  & \href{https://data.mendeley.com/datasets/4w6krnywhn/1}{MHCP2 [12]}
& Human & Colon  & 2  & 2  & 7101 & 100\%  & 97\%  & 24\% \\ \hline
\multirow{19}{*}{\begin{tabular}[c]{@{}c@{}}Intra-\\ Subject\end{tabular}} & \multirow{16}{*}{128} & 
\href{https://data.mendeley.com/datasets/svw96g68dv/4}{EHPCP2[5]}
& Human  & Prostate cancer  & 10  & 1 & 24465 & 100\%  & 92\%  & 37\% \\
& & \href{https://data.mendeley.com/datasets/svw96g68dv/4}{EHPCP1 [5]} 
& Human  & Prostate cancer  & 7 & 1   & 20987  & 100\%  & 92\%  & 34\% \\
& & \href{https://data.mendeley.com/datasets/4w6krnywhn/1}{MMBP2 [12]} 
& Mouse  & Brain & 4 & 1 & 17353 & 100\%  & 88\%  & 25\%  \\
 & & \href{https://data.mendeley.com/datasets/4w6krnywhn/1}{MMBP1 [12]}
 & Mouse & Brain & 4 & 1 & 17243 & 100\%  & 70\%  & 11\% \\
& & \href{https://data.mendeley.com/datasets/2bh5fchcv6/1}{AHSCC [1]}
& Human & \begin{tabular}[c]{@{}c@{}}Squamous \\ cell carcinoma\end{tabular}  & 4 & 1 & 10374 & 100\%  & 94\%  & 27\%
  \\
&  & \href{https://www.10xgenomics.com/resources/datasets/adult-human-brain-1-cerebral-cortex-unknown-orientation-stains-anti-gfap-anti-nfh-1-standard-1-1-0}{10XGHB [18]} 
& Human  & Brain  & 2 & 1  & 9882 & 100\%  & 89\%  & 23\% \\
& & \href{https://data.mendeley.com/datasets/nrbsxrk9mp/1}{FMBC [6]}
& Mouse & Brain - Coronal & 2 & 1 & 9132 & 100\%  & 80\%  & 24\%  \\
& & \href{https://www.10xgenomics.com/resources/datasets/human-breast-cancer-block-a-section-1-1-standard-1-0-0}{10GHBC [18]} 
& Human & Breast cancer & 2 & 1  & 7785 & 100\%  & 79\%  & 12\% \\
& & \href{https://data.mendeley.com/datasets/4w6krnywhn/1}{MMBO [12]}
& Mouse  & Bone & 4   & 1   & 7184 & 100\%  & 87\%  & 17\%   \\
& & \href{https://www.10xgenomics.com/resources/datasets/mouse-brain-serial-section-2-sagittal-posterior-1-standard-1-1-0}{10XGMBSP [18]}
& Mouse & \begin{tabular}[c]{@{}c@{}}Brain sagittal \\ posterior\end{tabular} & 2 & 1 & 6644 & 100\% & 79\% & 21\%  \\
& &  \href{https://data.mendeley.com/datasets/4w6krnywhn/1}{MHPBTP1 [12]} 
& Human & \begin{tabular}[c]{@{}c@{}}Pediatric brain \\ tumor\end{tabular} & 4 & 1 & 5937  & 100\% & 73\% & 21\%  \\
& &  \href{https://www.10xgenomics.com/resources/datasets/adult-mouse-brain-section-1-coronal-stains-dapi-anti-neu-n-1-standard-1-1-0}{10XGMBC [18]} 
& Mouse  & Brain coronal  & 2 & 1  & 5709 & 100\% & 80\% & 18\%   \\
&  &  \href{https://www.10xgenomics.com/resources/datasets/mouse-brain-serial-section-1-sagittal-anterior-1-standard-1-0-0}{10XGMBSA [18]} 
& Mouse  & \begin{tabular}[c]{@{}c@{}}Brain sagittal \\ anterior\end{tabular}  & 2 & 1  & 5520 & 100\% & 75\% & 15\%
  \\
&  &  \href{https://data.mendeley.com/datasets/nrbsxrk9mp/1}{FMOB [6]} 
& Mouse & Brain   & 2 & 1  & 2938 & 100\% & 67\% & 10\%  \\
&  & \href{https://data.mendeley.com/datasets/xjtv62ncwr/2}{VLO [21]} 
& Human & Lung organoids & 4  & 1  & 1832 & 100\% & 90\% & 26\%  \\
& &  \href{https://data.mendeley.com/datasets/xjtv62ncwr/1}{VKO [21]} 
& Human & Kidney organoids & 3  & 1  & 1355 & 100\% & 92\% & 33\%   \\ \cline{2-11} 
& \multirow{3}{*}{32}   & \href{https://figshare.scilifelab.se/articles/dataset/Spatial_Multimodal_Analysis_SMA_-_Spatial_Transcriptomics/22778920}{VHS [20]} 
& Human  & Striatium & 4  & 1 & 19033 & 100\% & 97\% & 30\%  \\
&  & \href{https://data.mendeley.com/datasets/4w6krnywhn/1}{MHPBTP2 [12]} 
& Human & \begin{tabular}[c]{@{}c@{}}Pediatric brain \\ tumor\end{tabular} & 2  & 1 & 3163 & 100\% & 97\% & 30\%   \\
& & \href{https://data.mendeley.com/datasets/4w6krnywhn/1}{MHCP1 [12]}  
& Human & Colon  & 2  & 1  & 2225 & 100\% & 90\% & 16\% \\ \hline
\end{tabular}}
\label{table:spared_details}
\end{table}

\subsection{Benchmark of Existing Gene Prediction Methods}

We use SpaRED to evaluate eight state-of-the-art expression profile prediction mds. Among these,  STNet \citep{he2020integrating} inputs individual patches into a fine-tuned DenseNet-121 with a linear layer for prediction. Additionally, STNet averages predictions across 8 symmetries of each patch to determine the final output. HisToGene \citep{pang2021leveraging} splits a WSI into patches that are processed by a Visual Transformer (ViT) model. The output is the genetic profile of the WSI. Hist2ST \citep{zeng2022spatial} divides the input histology image into multiple patches, which are processed by a Convolutional Neural Network (CNN) to extract 2D visual features. These learned features are then passed through a Transformer, enabling the model to capture global dependencies within the WSI. The output is then processed by a Graph Neural Network (GNN) to capture spatial dependencies between neighboring patches. Finally, the resulting representations are used to predict gene expression levels. BLEEP {citep{xie2023spatially} employs bi-modal contrastive learning to map image patches and expression profiles in a shared latent space, leveraging paired data to enhance representation learning. SEPAL \citep{mejia2023SEPAL} fine-tunes a ViT backbone and subsequently refines its predictions applying a GNN that processes a neighborhood graph for each patch. Additionally, SEPAL supervises expression changes relative to the mean expression in the training data instead of the absolute expression value, a strategy denoted ($\Delta$) prediction. EGN \citep{yang2023exemplar} applies exemplar-guided learning, a prediction strategy that bases its estimations on patches that are visually similar to the target patch within a latent space. This model integrates a ViT backbone with an Exemplar Bridging (EB) block, which dynamically improves feature representations using the most relevant exemplars. Building upon this approach, EGGN \citep{yang2024spatial} introduces an enhanced framework that, given a tissue slide image, encodes its windows into a feature space, retrieves exemplars from a reference dataset, constructs a graph, and dynamically predicts the gene expression of each window using an exemplar-guided graph network. Finally, HGGEP \citep{li2024gene} enhances feature extraction using a Gradient Enhancement Module (GEM). The latent features are then processed through a Convolutional Block Attention Module (CBAM) and a Vision Transformer (ViT), which leverage attention mechanisms to refine feature representations. A Hypergraph Association Module (HAM) further captures high-order associations by modeling global and local feature relationships based on spatial proximity.

Remarkably, Hist2ST and HGGEP rely on a graph-based approach that requires all spots in the WSI, resulting in high GPU memory consumption as the number of spots increases. Consequently, they can only run on 2 out of 26 SpaRED datasets. To evaluate these two models on the other 24 SpaRED datasets and thus enable a fair comparison with the other methods, we modify the model inputs by dividing the WSI into smaller sections (quarters, ninths, or sixteenths) and merging the predictions. To verify that this modification did not significantly impact the models' performance, we conducted an experiment comparing the results of both models when using the same dataset, once with the complete WSI as input and once with the WSI divided into four parts. We used the VLO dataset for this experiment, as it is one of the datasets where the full image can be used as input. The results showed that dividing the image into four parts leads to an increase in MSE of 2.85\% for Hist2ST and 0.002\% for HGGEP, indicating a slight decrease in performance. These findings confirm that our methodology does not significantly affect the capacity of the model.

Alongside these models, our comprehensive benchmark also includes the performance of three baseline methods: a ShuffleNet \citep{zhang2018shufflenet} architecture that finetunes an image encoder with low computational cost, a ViT-B encoder \citep{dosovitskiy2020image} that reflects the impact of fine-tuning a state-of-the-art backbone for this task, and a ViT-B+$\Delta$ approach as suggested by \cite{mejia2023SEPAL}. Moreover, we search for the optimal learning rate in every dataset. Then, with this value fixed, we explore two training scenarios: using raw data directly and SpaCKLE-completed data.

\section{Data Completion with Transformers}\label{Gene Completion with Transformers}

\begin{figure}[t]  
    \includegraphics[width=0.99\textwidth]{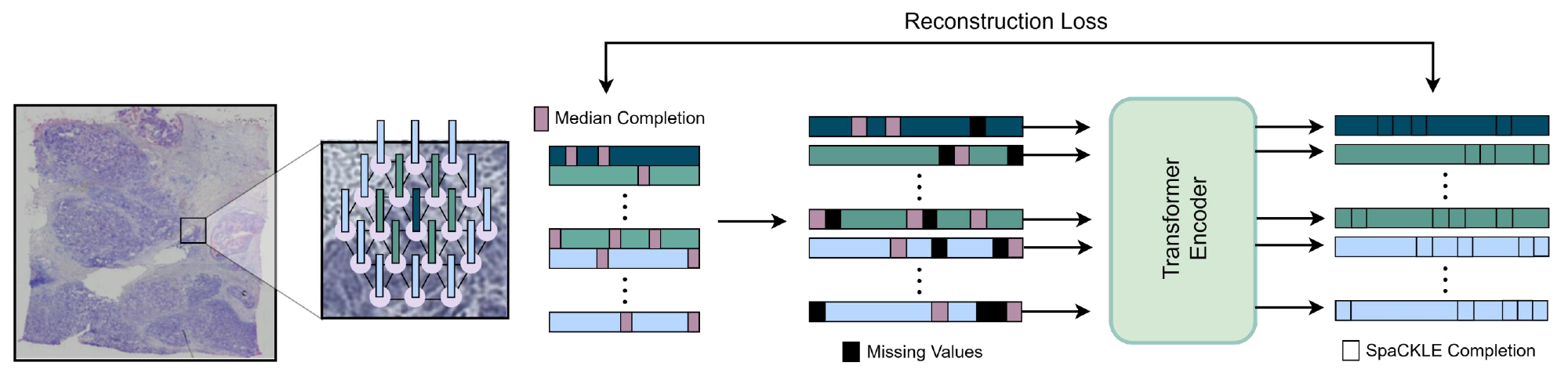}
    \caption{\textbf{SpaCKLE Overview.} Illustration of our data completion framework using a transformer-based model.}
    \label{fig:imputation}
\end{figure}

Inspired by the disruptive success of the transformer architecture for completion tasks such as language next token prediction \citep{vaswani2023attention} and visual reconstruction \citep{He2021}, we adapt these ideas to the ST domain. Fig. \ref{fig:imputation} illustrates SpaCKLE's training, a process that takes as a starting point data that we pre-completed using the median method proposed by \cite{mejia2023SEPAL}. This process ensures faster training convergence, guarantees non-zero predictions, and improves the overall performance of our completion model, as demonstrated in Section \ref{ablations}.

Given the median-completed expression vector $x \in \mathbb{R}^{g}$ of a particular spot $s$ with $g$ prediction genes and the expression matrix $V_x \in \mathbb{R}^{g~\times~n}$ that contains the genetic profile of the $n$ 2-hop neighbors in the Visium hexagonal geometry closest to $s$, we start by defining the expression matrix $E_x = \begin{bmatrix} x & V_x \end{bmatrix} \in \mathbb{R}^{g ~ \times ~(n+1)}$. Knowing the neighborhood of spots around $s$, we also define matrix $M_s \in \{0,1\}^{g ~\times ~ (n+1)}$, a binary mask that presents through 0 values the gene expressions within each spot that were originally missing in the dataset and pre-completed using median values. Moreover, to implement the masked-autoencoder-like workflow, we construct the matrix $M_{rand}(\rho) \in \{0,1\}^{g ~\times~ (n+1)}$, which randomly sets a fraction of $\rho=30\%$ values within the neighborhood as 1. The data points in $M_{rand}$ that have a value of 1 represent the elements in the neighborhood genetic profile that we set as candidates for our workflow to artificially hide. With these matrices, we then define the final random mask as

\begin{align}
    M_{mask} = M_s \odot M_{rand}.
\end{align}

$M_{mask}$ determines which values to hide in our input data by setting a fraction of its elements to 0, as follows:

\begin{align}
    E_m &= E_x \odot (1 - M_{mask}).
\end{align}

$M_{mask}$ is designed not to overlap with median-completed spots, guaranteeing that the ground truth used for computing evaluation metrics comes solely from values obtained with ST technology. After randomly masking $E_x$, we process it with a transformer encoder $T(\cdot)$ that leverages the self-attention mechanism:

\begin{align}
    \text{Attention}(Q, K, V) = \text{softmax}\left(\frac{QK^T}{\sqrt{d_k}}\right) V, 
\end{align}
to get a reconstructed version $\hat{E}_x$:
\begin{align}
    \hat{E}_x &= L_{out}\left(T(L_{in}(E_m))\right).
\end{align}

To accommodate different gene dimensionalities to a fixed transformer dimension $128$, we use the $L_{in}(\cdot)$ and $L_{out}(\cdot)$ linear adapters. We optimize an MSE loss between the two complete matrices:

\begin{align}
    \mathcal{L}=\left \| E_x - \hat{E}_x \right \|_2^2.
\label{eq:loss}
\end{align}

However, we only compute metrics and complete missing values using the masked elements from the first vector of the output. Hence, each component of the completed version $\hat{x}$ can be expressed as

\begin{align}
    \hat{x}_i = 
    \begin{cases} 
      x_i, & M_{mask}[i, 1]=0 \\
      \hat{E}_x[i, 1], & M_{mask}[i, 1]=1.
   \end{cases}
\end{align}

To reduce any potential bias when evaluating SpaCKLE on artificially hidden values, we perform a total of 10 assays in each testing process, where each assay involves a different random mask $M_{rand}$. Once we compute the evaluation metrics for each assay, we report their average value as the final result. On the other hand, during inference, we do not include $M_{rand}$ when defining $M_{mask}$ but rather only consider the originally missing values set as 0 in $M_s$ to remove the pre-completed values from the input data. After processing the input $E_m$ with SpaCKLE, we get a refined version of the gene expression profiles.

\subsection{Implementation Details:}

We train all our models on a NVIDIA Quadro RTX 8000 with a batch size of 256 and use an Adam \citep{kingma2017adam} optimizer with default PyTorch library parameters. We train one completion model for each dataset, and optimize each one using a range of ten different learning rates sampled on a logarithmic scale between $1 \times 10^{-5}$ and $1 \times 10^{-2}$. We also conduct this learning rate optimization on the gene expression prediction models of our benchmark. Furthermore, we use a constant learning rate during training, and to ensure the reproducibility of our experiments, we fix the random seed to 42 and set all relevant random number generators accordingly.

We handle both regression and completion problems as multivariate regression tasks and evaluate them using MSE and PCC. To select the best model, we save the one with the lowest validation MSE after 1,000 and 10,000 iterations for prediction and completion, respectively. All metrics are computed exclusively on real data for both the completion and the prediction task.

\section{Results and Discussion}

\subsection{Gene Completion Evaluation}

\begin{figure*}[t]  
    \includegraphics[width=0.99\textwidth]{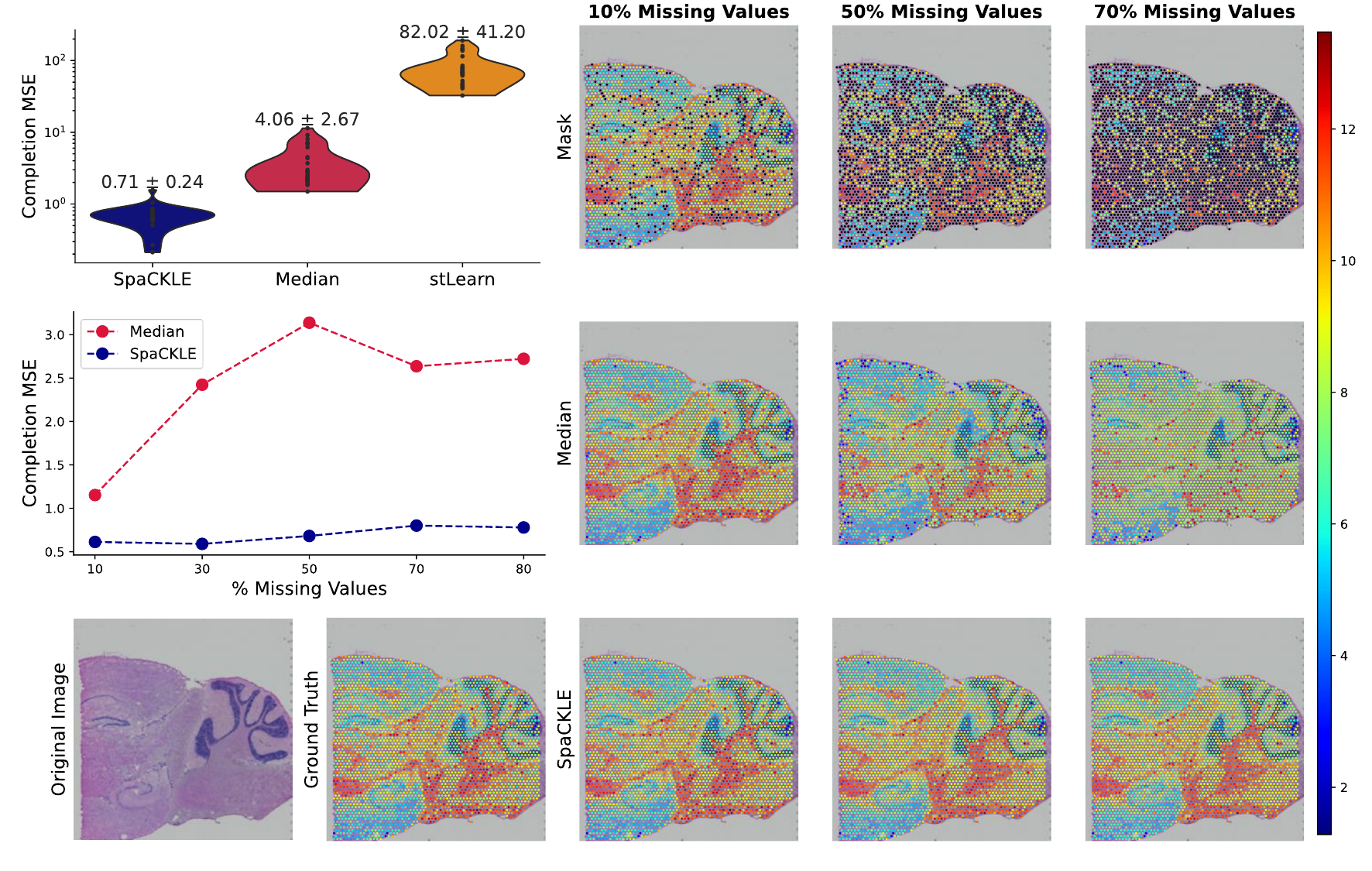}
    \caption{\textbf{Completion Methods Comparison.} Violin plot displaying completion MSE scores for each method (SpaCKLE, Median and stLearn) across all datasets in SpaRED (upper left). Line plot displaying completion MSE for the median and SpaCKLE methods across different percentages of synthetically masked data (middle left). Qualitative results showing gene completion for increasing synthetic masking percentages (row 1) with the median method (row 2) and SpaCKLE (row 3).}
    \label{fig:imputation_2}
\end{figure*}

The violin plot in Fig. \ref{fig:imputation_2} presents a comparison of the logarithmic MSE for data completion using SpaCKLE, the median completion method, and stLearn across SpaRED. The results indicate that SpaCKLE outperforms alternative completion methods, with a relative 82.5\% MSE reduction compared to the median method and by two orders of magnitude concerning stLearn. Notably, stLearn presents the highest MSE in the entirety of SpaRED, which conveys its inability to restore masked data. These results are consistent with those reported in \citep{avsar2023comparative}, where stLearn's completion predictions included a high proportion of zero values. It is noteworthy that the median method is based solely on the adjacent expression of a single gene, an approach that, although straightforward, does not consider the broader genetic context. In contrast, SpaCKLE has access to the complete genetic profile of the neighboring spots. Thus, we hypothesize that our transformer architecture is leveraging the full expression profile of the empty spot's vicinity to enhance completion predictions.

To thoroughly assess the robustness of our approach, we characterize the completion performance when synthetically corrupting increasing percentages of data in the 10XGMBSP dataset. The line graph in Fig. \ref{fig:imputation_2} show how the completion's accuracy changes for the median and SpaCKLE methods with various masking percentages. For visualization purposes, we only display MSE results for SpaCKLE and the median method since stLearn has a significantly higher MSE. We observe that, as the task gets more challenging with a greater percentage of missing data, SpaCKLE outperforms the median completion method by a larger margin. Specifically, while SpaCKLE's MSE for data completion shows a slight increase with more missing values, the MSE for the median method rises dramatically, growing from a minimum of 1.2 to over 3 as the percentage of missing values increases. This demonstrates SpaCKLE's superior ability to handle larger amounts of missing data effectively. The predicted expression maps support these observations, showing that SpaCKLE strongly approximates the ground truth patterns even at a missing value percentage of 70\%. Conversely, the uniformity in the color pattern of the predictions from the median method demonstrates that this strategy repeatedly imposes the global median when it cannot find a local value due to the high fraction of missing data. This behavior impairs the expression profiles by homogenizing the gene's activity in the tissue and removing spatial information. 

\begin{figure*}[t]  

    \includegraphics[width=0.99\textwidth]{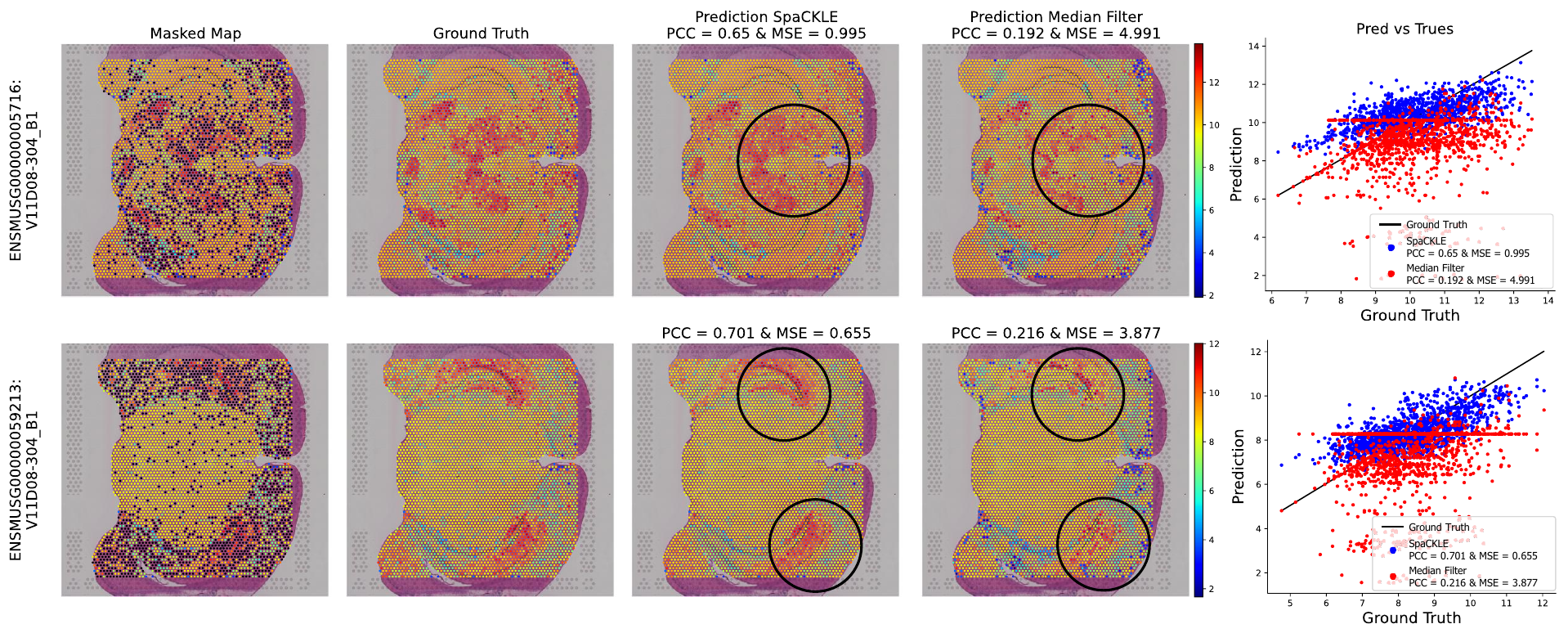}
    \caption{\textbf{Gene Completion Results.} Qualitative results showing gene completion at a 30\% masking percentage (column 1). Column 2 includes the real values, while column 3 displays results from SpaCKLE and column 4 shows results from the median method. The scatter plot in column 5 compares the predicted expression values to the actual ground truth values for all spots of a specific gene. Blue dots represent the outcomes from SpaCKLE, and red dots indicate the results from the median filter.}
    \label{fig:mask_30}

\end{figure*}

Fig. \ref{fig:mask_30} presents additional examples of expression predictions made by SpaCKLE compared to those made by the median method for cases with 30\% artificial missing data. The scatter plot visualizes the expression predictions against the ground truth across all the masked spots for a specific gene. Blue dots represent the predictions from SpaCKLE, while red dots correspond to the predictions from the median method. The black diagonal line indicates the ideal scenario in which the predictions perfectly match the ground truth. The results indicate that the blue dots follow the black line more closely compared to the red dots, suggesting that SpaCKLE has a greater capacity to accurately predict gene expression across most spots. The red dots, which represent the median method predictions, show a significant percentage of spots arranged in a straight line. This pattern indicates that many spots are assigned the same expression value. This observation supports the earlier analysis, which highlighted that the median method tends to predict a uniform value -corresponding to the global median- when it is unable to determine the local median. 

The qualitative results highlight the advantage of using SpaCKLE for data completion over the median filter, particularly in preserving specific patterns across different regions. The black-circled areas indicate sections where the median filter struggles to recover the true expression values of a given gene. These regions often correspond to areas with clustered missing values, which is expected since the median filter relies solely on adjacent gene expression. As the proportion of missing values in a spot’s vicinity increases, the median method becomes less effective at making accurate predictions. In contrast, SpaCKLE leverages self-attention to predict gene expression more accurately by incorporating information from the full expression profile of surrounding spots.

\begin{table}[t]
\centering
\caption{Comparison of SpaCKLE and three ablated configurations: (i) using Context Genes, (ii) incorporating Visual Features, and (iii) training without data pre-completion. Metrics reported are average MSE and PCC on all SpaRED datasets. Best configuration is bolded. }
\begin{tabular}{@{}lll@{}}
\toprule
 & \textbf{  MSE  } & \textbf{  PCC  } \\ \midrule
SpaCKLE & \textbf{0.713} & \textbf{0.600} \\
SpaCKLE with Context Genes & 0.787 & 0.534 \\ 
SpaCKLE with Visual Features & 0.854 & 0.533 \\
SpaCKLE without data pre-completion & 6.166 & 0.165 \\
\bottomrule
\end{tabular}
\label{tab:ablation}
\end{table}

\begin{table}[t]
\centering
\caption{Effect of spatial neighborhood size on SpaCKLE’s completion metrics. Results for 0, 6, 18, and 36 neighbors (0–3 Visium hops) are shown, including average MSE and PCC on all SpaRED datasets. Best configuration is bolded.}
\begin{tabular}{@{}lcccc@{}}
\toprule
& \multicolumn{4}{c}{\textbf{Number of Neighbors}} \\ \midrule
& \textbf{0} & \textbf{6} & \textbf{18} & \textbf{36} \\
\textbf{MSE} & 0.8136  & 0.7263  & \textbf{0.7134}      & 0.7156   \\
\textbf{PCC} & 0.4727     & 0.5290     & 0.5316      & \textbf{0.5321}      \\ \bottomrule
\end{tabular}
\label{tab:neighbors}
\end{table}

\subsubsection{Ablation Experiments}
\label{ablations}

To understand the contributions of each component in SpaCKLE, we carry out a series of controlled ablations on all SpaRED datasets. First, we assess the impact of our pre-completion step. In the full pipeline, we complete missing gene expression values with the median-completion method of \cite{mejia2023SEPAL} before training. In the ablated variant, we train directly on the raw data with missing entries. As shown in Table \ref{tab:ablation}, median pre-completion reduces the mean squared error (MSE) by a factor of eight and increases the Pearson correlation coefficient (PCC) by nearly threefold, confirming that filling missing spots with a simple median estimate provides richer signals for learning and leads to substantially higher‐quality reconstructions. To prevent the model from memorizing medians, we mask out only original, non-precompleted values during synthetic masking in training.

Next, we explore the use of visual information from the ST spots in the input neighborhood. In this case, the model's workflow receives a matrix $H_x \in \mathbb{R}^{d ~\times~ (n+1)}$ along with matrix $E_m$, described in detail in section \ref{Gene Completion with Transformers}. Matrix $H_x$ contains the image embeddings of dimension $d$ of the $n+1$ spots in the incoming neighborhood, which we obtain by processing their ST patches with a ViT model backbone that we fine-tuned for gene expression prediction, as proposed by \cite{mejia2023SEPAL}. We concatenate the visual features with the genetic profile of its corresponding spot and feed this combined representation into the transformer encoder as part of SpaCKLE's framework.

In contrast to our original assumptions, the inclusion of visual features leads to a $\sim$20\% increase in MSE and an $\sim$11\% drop in PCC, indicating a reduction in completion performance (Table~\ref{tab:ablation}). These results prompt further analysis, and we suspect that several factors may explain this behavior. First, although we fine-tuned the ViT model for gene expression prediction, we freeze its weights during SpaCKLE’s training. This may have limited the capacity of the visual features to adapt to the masked reconstruction objective of our transformer-based framework. Additionally, we hypothesize that the use of domain-specific histology foundation models, such as UNI (\cite{uni-encoder}), could yield more task-relevant visual representations. Finally, another possible explanation is that the direct concatenation of gene expression vectors and visual embeddings, which are two modalities with inherently different distributions and scales, introduces imbalances that negatively impact learning. Our findings suggest that more sophisticated fusion mechanisms, such as modality-aware normalization, gated integration, or joint end-to-end training, may be beneficial for fully leveraging histological information in gene expression completion.

We also investigate the effect of profile length. While SpaRED defaults to selecting 32 or 128 genes by Moran’s I score, we extend this to 1024 genes, defining $x'\in \mathbb{R}^{1024}$, \(E_{x'} = \begin{bmatrix} x^{'} & V_{x'} \end{bmatrix} \in \mathbb{R}^{1024 ~\times~ (n+1)}\), where the extra genes have the next highest spatial autocorrelation. We mask only the original 32/128 genes using a random mask $M'_{rand} \in \{0,1\}^{1024 ~\times~ (n+1)}$ that ensures new genes remain zero for treating them purely as context. This wider context increases MSE by 10.4\% and drops PCC by 11.1\%, suggesting that genes with weaker spatial patterns add noise rather than helpful cues.

Finally, we examine how the number of spatial neighbors influences completion. We vary the neighborhood size among 0, 6, 18 and 36 spots - corresponding to 0, 1, 2, or 3 Visium hops, respectively - and retrained SpaCKLE with the same masking scheme. As Table \ref{tab:neighbors} shows, the jump from 0 to 6 neighbors yields the largest improvement, expanding to two hops (18 neighbors) delivers a modest further improvement, but pushing to three hops (36 neighbors) produces a 0.3\% increase in MSE and only 0.1\% improvement in PCC. Given this plateau and the computational cost of larger neighborhoods, we select 18 neighbors as the optimal number of neighbors.

Together, these ablations confirm our design choices: (1) median pre-completion is essential for strong recovery, (2) a moderate neighborhood of 18 spots provides the best trade-off between performance and efficiency, and (3) neither adding histology embeddings nor context genes yields consistent benefit for SpaCKLE’s core completion task.

\subsection{Gene Prediction Benchmark}

\begin{figure*}[t]
    \includegraphics[width=0.99\textwidth]{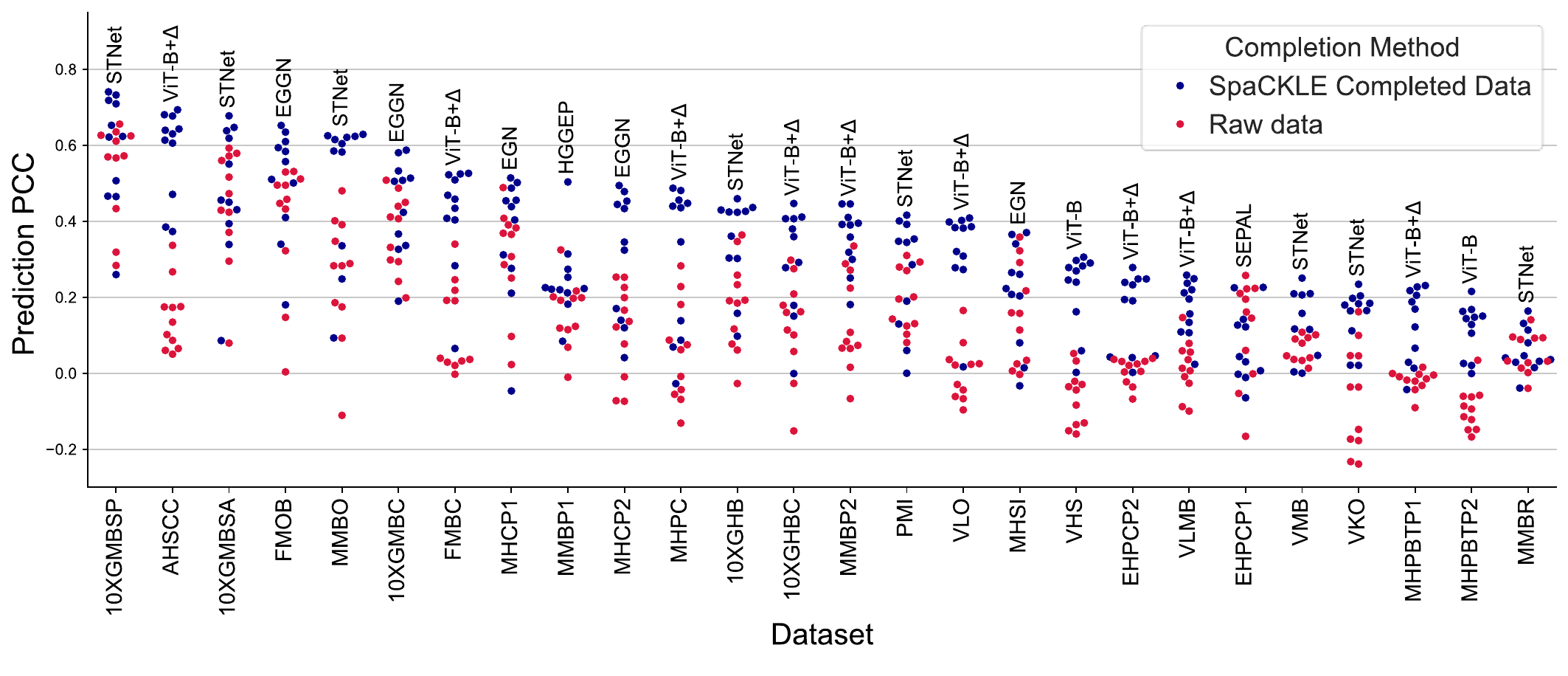}
    \caption{\textbf{Impact of SpaCKLE on SpaRED Benchmark.} Prediction Pearson Correlation Coefficient for each model across all the datasets in SpaRED. For each dataset, the state-of-the-art model that obtains the highest Pearson Correlation Coefficient is included. As evidenced by the red (raw data) and blue dots (SpaCKLE-completed data), SpaCKLE improves performance across all methods in every dataset.}
    \label{fig:general_results}
\end{figure*}

Fig. \ref{fig:general_results} shows the performance of all methods for every dataset when trained under our two scenarios (raw and SpaCKLE-completed data). It is clear that the prediction performance significantly improves when applying SpaCKLE to every dataset and, in some cases, the best PCC increases to 0.36 points (AHSCC). This result pinpoints the importance of acknowledging missing data for the prediction task and proves the significance of including gene completion in ST pipelines.

Comparing datasets' difficulty, we find that the most challenging dataset to predict was MMBR (PCC=0.16), while 10XGMBSP emerged as the least difficult (PCC=0.74). When inspecting each dataset's characteristics, we observe that the organism does not appear to have a significant impact on the difficulty of the task, as the mean prediction PCC achieved for mouse and human datasets is very similar. Furthermore, a larger number of available genes (due to better quality) facilitates prediction, which is evident in a higher average and maximum performance on the datasets with 128 genes compared to those with 32 genes. Finally, results also demonstrate that generalizing in an intra-subject manner typically makes the prediction easier than inter-subject (See Fig.\ref{fig:tissue_type_analysis}. a). 

We analyze the prediction performance across various tissue types. Fig. \ref{fig:tissue_type_analysis}. b presents the performance of each dataset in SpaRED, categorized by tissue type, where the bars indicate the average PCC for each type of tissue. On average, the best prediction performance was observed for skin tissue, while the lowest was for kidney tissue. However, the distribution of tissue types in SpaRED is highly imbalanced, with some tissue types being underrepresented. Notably, both skin and kidney tissues are represented by only a single dataset, making it unreliable to draw definitive conclusions about whether certain tissues are inherently easier to predict than others. The observed differences may be influenced by dataset-specific characteristics rather than general tissue properties.

\begin{figure}[t]
    \centering
    \includegraphics[width=0.84\textwidth]{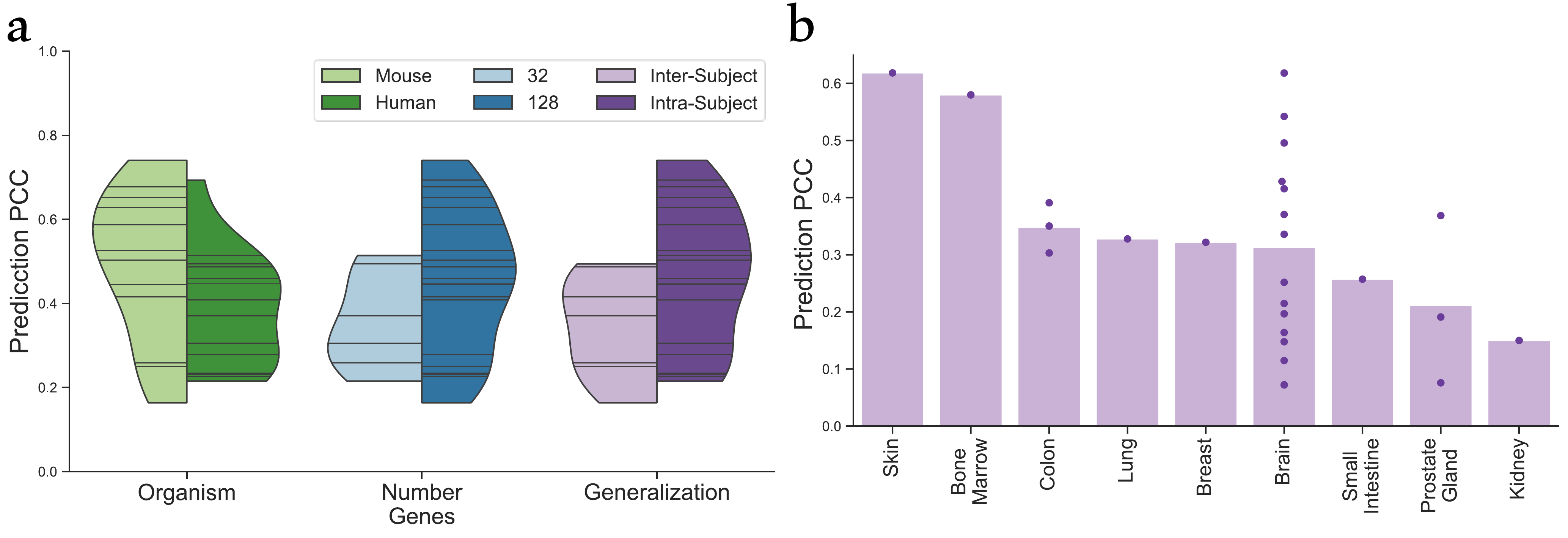}
    \caption{\textbf{Effect of SpaRED Categories on Benchmark.} (a) Violin plots illustrate the variation in key characteristics of the datasets, such as organism, number of genes, and generalization task. The data depicted represent the best prediction PCC achieved for each dataset within the SpaRED collection. (b) Bar charts display the prediction performance, measured by PCC, for each type of tissue analyzed in SpaRED. The dots represent the best prediction PCC achieved for each dataset, while the bars indicate the average PCC across each type of tissue.}
    \label{fig:tissue_type_analysis}
\end{figure}

\begin{table}[]
\caption{The matrix illustrates the statistically significant differences in MSE among all models across all datasets. A Dunn test with a 5\% significance level was used to identify these differences. Differences that are statistically significant between models are highlighted in \textbf{bold}.}
\resizebox{0.9\textwidth}{!}{
{\fontsize{7}{10}\selectfont
\begin{tabular}{l|ccccccccccc}
\hline
 & \multicolumn{1}{l}{\textbf{Hist2ST}} 
& \multicolumn{1}{l}{\textbf{HGGEP}} 
& \multicolumn{1}{l}{\textbf{ShuffleNet}} & \multicolumn{1}{l}{\textbf{STNet}} & \multicolumn{1}{l}{\textbf{EGN}} & \multicolumn{1}{l}{\textbf{EGGN}} & \multicolumn{1}{l}{\textbf{HisToGene}} & \multicolumn{1}{l}{\textbf{ViT-B}} & \multicolumn{1}{l}{\textbf{VIT-B+$\Delta$}} & \multicolumn{1}{l}{\textbf{SEPAL}}
& \multicolumn{1}{l}{\textbf{BLEEP}} \\ \hline

\textbf{Hist2ST} & - & 1.0 & 1.0 & 0.178 & 1.0 & 1.0 & 0.068 & 1.0 & \textbf{0.029} & 0.191 & 0.852 \\

\textbf{HGGEP} & - & - & {$\mathbf{8.782 \times 10 ^{-3}}$} & {$\mathbf{1.513 \times 10 ^{-5}}$} & \textbf{0.047} &{$\mathbf{5.495 \times 10 ^{-3}}$} & 1.0 &{$\mathbf{1.188 \times 10 ^{-3}}$} & {$\mathbf{8.208 \times 10^{-7}}$} & {$\mathbf{1.691 \times 10 ^{-5}}$} & 1.0 \\

\textbf{ShuffleNet} & - & - & - & 1.0 & 1.0 & 1.0 & {$\mathbf{9.039 \times 10^{-5}}$} & 1.0 & 1.0 & 1.0 & {$\mathbf{3.480 \times 10^{-3}}$} \\

\textbf{STNet} & - & - & - & - & 1.0 & 1.0 & {$\mathbf{4.432 \times 10^{-8}}$} & 1.0 & 1.0 & 1.0 & {$\mathbf{4.464 \times 10^{-6}}$} \\

\textbf{EGN} & - & - & - & - & - & 1.0 &{$\mathbf{7.359 \times 10^{-4}}$} & 1.0 & 1.0 & 1.0 & \textbf{0.021} \\

\textbf{EGGN} & - & - & - & - & - & - & {$\mathbf{5.083 \times 10^{-5}}$} & 1.0 & 1.0 & 1.0 & {$\mathbf{2.124 \times 10^{-3}}$} \\

\textbf{HisToGene} & - & - & - & - & - & - & - & {$\mathbf{7.903 \times 10^{-6}}$} & {$\mathbf{1.493 \times 10^{-9}}$} & {$\mathbf{5.052 \times 10^{-8}}$} & 1.0 \\

\textbf{ViT-B} & - & - & - & - & - & - & - & - & 1.0 & 1.0 & {$\mathbf{4.254 \times 10^{-4}}$} \\

\textbf{ViT-B+$\Delta$} & - & - & - & - & - & - & - & - & - & 1.0 & {$\mathbf{2.161 \times 10^{-7}}$} \\

\textbf{SEPAL} & - & - & - & - & - & - & - & - & - & - & {$\mathbf{5.015 \times 10^{-6}}$} \\

\textbf{BLEEP} & - & - & - & - & - & - & - & - & - & - & - \\

\hline
\end{tabular}}}
\label{table:statistical_analysis}
\end{table}

We display the results of evaluating the 8 state-of-the-art models on SpaRED, as well as the baseline experiments on Fig. \ref{fig:benchmark}.a sorted by best average performance. The normalized MSE metric indicates how close every model's results are to the best performance achieved on each dataset. Results show that ViT-B+$\Delta$ attains the best gene expression predictions on average, despite being one of the most straightforward approaches for the prediction task. Moreover, the pie chart showcases that STNet and ViT-B+$\Delta$ emerge most frequently as the best methods. Interestingly, we notice that SEPAL, which is built on top of ViT-B+$\Delta$, falls behind the latter. This contrast reveals that incorporating local vicinity information does not necessarily improve the outputs and that focusing on predicting the $\Delta$ from the mean expression is already a powerful strategy. 

Table \ref{table:statistical_analysis} illustrates the statistical differences in MSE performance across all datasets. A Dunn test with a 5\% significance level reveals that most methods do not exhibit statistically significant differences in performance. While previous results indicate that ViT-B+$\Delta$ and STNet most frequently achieve the best results, with ViT-B+$\Delta$ obtaining the highest average performance, the statistical analysis confirms that these improvements are not significant when compared to most other methods. Notably, ViT-B+$\Delta$ shows a statistically significant advantage only over Hist2ST, HGGEP, HisToGene, and BLEEP, while no significant advantage is observed over the other methods. These results indicate that no single state-of-the-art method is definitively superior, highlighting that the existing strategies for improving gene expression prediction remain insufficient. This underscores the need for novel approaches to enhance ST-related tasks.

\begin{figure*}[] 
    \includegraphics[width=0.99\textwidth]{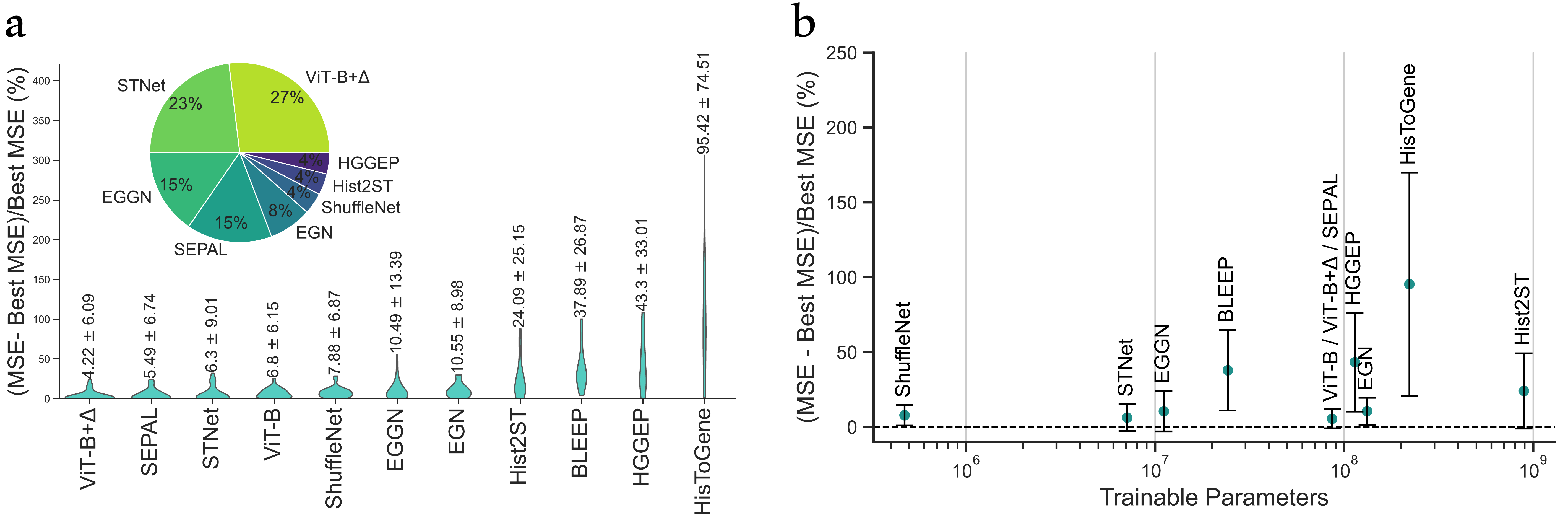}
    \caption{\textbf{SpaRED Benchmark Results.} (a) Violin plot: normalized prediction MSE of each model across all datasets within SpaRED, with normalization done against the best MSE obtained on each dataset. The mean and standard deviation of the methods are included at the top of each violin. Pie chart: percentage of datasets within SpaRED for which each model achieves the best prediction MSE. (b) Mean normalized prediction MSE against the number of trainable parameters for each model.} 
    \label{fig:benchmark}
\end{figure*}

Our results also indicate that more complex architectures do not necessarily provide superior predictions on our benchmark. This behavior is also supported by Fig. \ref{fig:benchmark}.b, where Hist2ST ranks as the method with the most trainable parameters but performs worse than methods with orders of magnitude fewer parameters. In contrast, ShuffleNet is the method with the fewest parameters and offers a competitive performance. We hypothesize that this counterintuitive trend is caused by the limited scale of publicly available datasets (the biggest SpaRED dataset contains 43,804 spots), probably leading to overfitting in bigger models.

\section{Conclusions}

In this paper, we present SpaRED, a systematically curated Visium database comprising 26 standardized datasets that emerges as a novel standard point of comparison for gene expression prediction from histology images methods. We also introduce SpaCKLE, a transformer-based model that successfully overcomes the dropout limitations in ST technology, completing gene expression values even when the missing data fraction is up to 70\%. SpaCKLE achieves an 82.5\% reduction in MSE compared to the median-based imputation method, significantly improving the quality of gene expression completion. Moreover, our benchmarking of eight state-of-the-art models on SpaRED demonstrates that integrating SpaCKLE as a preprocessing step enhances prediction performance across all methods. However, statistical analysis reveals that most methods do not exhibit significant performance differences when trained on the same data, suggesting that existing approaches for robust gene expression prediction remain insufficient. Furthermore, our results highlight that increasing model complexity does not necessarily lead to better gene expression predictions, emphasizing the need for novel strategies to advance ST. Consequently, our work represents a significant advancement in the automation of ST and is intended to promote further research in this field.

\section*{Acknowledgments} 
Gabriel M. Mejia and Daniela Vega acknowledge the support of UniAndes-GoogleDeepMind Scholarships 2022 and 2024 respectively. This work was supported by Azure sponsorship credits granted by Microsoft’s AI for Good Research Lab.

\section*{Declaration of Generative AI and AI-assisted technologies in the writing process}
During the preparation of this work, the authors used Grammarly Premium and Open AI's ChatGPT in order to assist in drafting, checking spelling and grammar, organizing text, and paraphrasing or looking for alternative vocabulary. After using this tool/service, the authors reviewed and edited the content as needed and take full responsibility for the content of the publication.

\printcredits

\section*{Data Availability}
Data is published at \url{https://github.com/BCV-Uniandes/SpaRED}, \url{https://drive.google.com/drive/folders/15W_rZlt5PhUlslM-u5_jw9etjkGRXb-N}

\bibliographystyle{cas-model2-names}
\bibliography{references}

\begin{thebibliography}{44}
\expandafter\ifx\csname natexlab\endcsname\relax\def\natexlab#1{#1}\fi
\providecommand{\url}[1]{\texttt{#1}}
\providecommand{\href}[2]{#2}
\providecommand{\path}[1]{#1}
\providecommand{\DOIprefix}{doi:}
\providecommand{\ArXivprefix}{arXiv:}
\providecommand{\URLprefix}{URL: }
\providecommand{\Pubmedprefix}{pmid:}
\providecommand{\doi}[1]{\href{http://dx.doi.org/#1}{\path{#1}}}
\providecommand{\Pubmed}[1]{\href{pmid:#1}{\path{#1}}}
\providecommand{\bibinfo}[2]{#2}
\ifx\xfnm\relax \def\xfnm[#1]{\unskip,\space#1}\fi
\bibitem[{Abalo et~al.(2021)Abalo, Thrane, Ji et~al.}]{Abalo2021}
\bibinfo{author}{Abalo, X.}, \bibinfo{author}{Thrane, K.}, \bibinfo{author}{Ji, A.L.}, et~al., \bibinfo{year}{2021}.
\newblock \bibinfo{title}{Human squamous cell carcinoma, visium} \bibinfo{volume}{1}.
\newblock \DOIprefix\doi{10.17632/2bh5fchcv6.1}.
\bibitem[{Abdelaal et~al.(2020)Abdelaal, Mourragui, Mahfouz and Reinders}]{abdelaal2020spage}
\bibinfo{author}{Abdelaal, T.}, \bibinfo{author}{Mourragui, S.}, \bibinfo{author}{Mahfouz, A.}, \bibinfo{author}{Reinders, M.J.}, \bibinfo{year}{2020}.
\newblock \bibinfo{title}{Spage: spatial gene enhancement using scrna-seq}.
\newblock \bibinfo{journal}{Nucleic acids research} \bibinfo{volume}{48}, \bibinfo{pages}{e107--e107}.
\bibitem[{Avşar and Pir(2023)}]{avsar2023comparative}
\bibinfo{author}{Avşar, G.}, \bibinfo{author}{Pir, P.}, \bibinfo{year}{2023}.
\newblock \bibinfo{title}{A comparative performance evaluation of imputation methods in spatially resolved transcriptomics data}.
\newblock \bibinfo{journal}{Molecular Omics} \bibinfo{volume}{19}, \bibinfo{pages}{162–173}.
\newblock \DOIprefix\doi{10.1039/d2mo00266c}.
\bibitem[{Biancalani et~al.(2021)Biancalani, Scalia, Buffoni et~al.}]{biancalani2021deep}
\bibinfo{author}{Biancalani, T.}, \bibinfo{author}{Scalia, G.}, \bibinfo{author}{Buffoni, L.}, et~al., \bibinfo{year}{2021}.
\newblock \bibinfo{title}{Deep learning and alignment of spatially resolved single-cell transcriptomes with tangram}.
\newblock \bibinfo{journal}{Nature methods} \bibinfo{volume}{18}, \bibinfo{pages}{1352--1362}.
\bibitem[{Chen et~al.(2024a)Chen, Zhou, Wu, Zhang, Li and Li}]{chen2024stimage}
\bibinfo{author}{Chen, J.}, \bibinfo{author}{Zhou, M.}, \bibinfo{author}{Wu, W.}, \bibinfo{author}{Zhang, J.}, \bibinfo{author}{Li, Y.}, \bibinfo{author}{Li, D.}, \bibinfo{year}{2024}a.
\newblock \bibinfo{title}{Stimage-1k4m: A histopathology image-gene expression dataset for spatial transcriptomics}.
\newblock \bibinfo{journal}{ArXiv} , \bibinfo{pages}{arXiv--2406}.
\bibitem[{Chen et~al.(2015)Chen, Boettiger, Moffitt et~al.}]{merfish}
\bibinfo{author}{Chen, K.H.}, \bibinfo{author}{Boettiger, A.N.}, \bibinfo{author}{Moffitt, J.R.}, et~al., \bibinfo{year}{2015}.
\newblock \bibinfo{title}{Spatially resolved, highly multiplexed rna profiling in single cells}.
\newblock \bibinfo{journal}{Science} \bibinfo{volume}{348}.
\newblock \URLprefix \url{https://www.science.org/doi/10.1126/science.aaa6090}, \DOIprefix\doi{10.1126/science.aaa6090}.
\bibitem[{Chen et~al.(2024b)Chen, Ding, Lu, Williamson, Jaume, Song, Chen, Zhang, Shao, Shaban et~al.}]{uni-encoder}
\bibinfo{author}{Chen, R.J.}, \bibinfo{author}{Ding, T.}, \bibinfo{author}{Lu, M.Y.}, \bibinfo{author}{Williamson, D.F.}, \bibinfo{author}{Jaume, G.}, \bibinfo{author}{Song, A.H.}, \bibinfo{author}{Chen, B.}, \bibinfo{author}{Zhang, A.}, \bibinfo{author}{Shao, D.}, \bibinfo{author}{Shaban, M.}, et~al., \bibinfo{year}{2024}b.
\newblock \bibinfo{title}{Towards a general-purpose foundation model for computational pathology}.
\newblock \bibinfo{journal}{Nature Medicine} \bibinfo{volume}{30}, \bibinfo{pages}{850--862}.
\bibitem[{Choe et~al.(2023)Choe, Pak, Pang, Hao and Yang}]{choe2023advances}
\bibinfo{author}{Choe, K.}, \bibinfo{author}{Pak, U.}, \bibinfo{author}{Pang, Y.}, \bibinfo{author}{Hao, W.}, \bibinfo{author}{Yang, X.}, \bibinfo{year}{2023}.
\newblock \bibinfo{title}{Advances and challenges in spatial transcriptomics for developmental biology}.
\newblock \bibinfo{journal}{Biomolecules} \bibinfo{volume}{13}, \bibinfo{pages}{156}.
\bibitem[{Dosovitskiy et~al.(2020)Dosovitskiy, Beyer, Kolesnikov et~al.}]{dosovitskiy2020image}
\bibinfo{author}{Dosovitskiy, A.}, \bibinfo{author}{Beyer, L.}, \bibinfo{author}{Kolesnikov, A.}, et~al., \bibinfo{year}{2020}.
\newblock \bibinfo{title}{An image is worth 16x16 words: Transformers for image recognition at scale}.
\newblock \bibinfo{journal}{arXiv preprint arXiv:2010.11929} .
\bibitem[{Erickson et~al.(2022)Erickson, He, Berglund et~al.}]{erickson2022spatially}
\bibinfo{author}{Erickson, A.}, \bibinfo{author}{He, M.}, \bibinfo{author}{Berglund, E.}, et~al., \bibinfo{year}{2022}.
\newblock \bibinfo{title}{Spatially resolved clonal copy number alterations in benign and malignant tissue}.
\newblock \bibinfo{journal}{Nature} \bibinfo{volume}{608}, \bibinfo{pages}{360--367}.
\bibitem[{Fan et~al.(2023)Fan, Andrusivov{\'a}, Wu et~al.}]{fan2023expansion}
\bibinfo{author}{Fan, Y.}, \bibinfo{author}{Andrusivov{\'a}, {\v{Z}}.}, \bibinfo{author}{Wu, Y.}, et~al., \bibinfo{year}{2023}.
\newblock \bibinfo{title}{Expansion spatial transcriptomics}.
\newblock \bibinfo{journal}{Nature Methods} , \bibinfo{pages}{1--4}.
\bibitem[{He et~al.(2020)He, Bergenstr{\aa}hle, Stenbeck et~al.}]{he2020integrating}
\bibinfo{author}{He, B.}, \bibinfo{author}{Bergenstr{\aa}hle, L.}, \bibinfo{author}{Stenbeck, L.}, et~al., \bibinfo{year}{2020}.
\newblock \bibinfo{title}{Integrating spatial gene expression and breast tumour morphology via deep learning}.
\newblock \bibinfo{journal}{Nature biomedical engineering} \bibinfo{volume}{4}, \bibinfo{pages}{827--834}.
\bibitem[{He et~al.(2021)He, Chen, Xie, Li, Dollar and Girshick}]{He2021}
\bibinfo{author}{He, K.}, \bibinfo{author}{Chen, X.}, \bibinfo{author}{Xie, S.}, \bibinfo{author}{Li, Y.}, \bibinfo{author}{Dollar, P.}, \bibinfo{author}{Girshick, R.}, \bibinfo{year}{2021}.
\newblock \bibinfo{title}{Masked autoencoders are scalable vision learners}.
\newblock \bibinfo{journal}{Proceedings of the IEEE Computer Society Conference on Computer Vision and Pattern Recognition} \bibinfo{volume}{2022-June}, \bibinfo{pages}{15979--15988}.
\newblock \URLprefix \url{https://arxiv.org/abs/2111.06377v3}, \DOIprefix\doi{10.1109/CVPR52688.2022.01553}.
\bibitem[{Jaume et~al.(2024)Jaume, Doucet, Song, Lu, Almagro~P{\'e}rez, Wagner, Vaidya, Chen, Williamson, Kim et~al.}]{hest1k}
\bibinfo{author}{Jaume, G.}, \bibinfo{author}{Doucet, P.}, \bibinfo{author}{Song, A.}, \bibinfo{author}{Lu, M.Y.}, \bibinfo{author}{Almagro~P{\'e}rez, C.}, \bibinfo{author}{Wagner, S.}, \bibinfo{author}{Vaidya, A.}, \bibinfo{author}{Chen, R.}, \bibinfo{author}{Williamson, D.}, \bibinfo{author}{Kim, A.}, et~al., \bibinfo{year}{2024}.
\newblock \bibinfo{title}{Hest-1k: A dataset for spatial transcriptomics and histology image analysis}.
\newblock \bibinfo{journal}{Advances in Neural Information Processing Systems} \bibinfo{volume}{37}, \bibinfo{pages}{53798--53833}.
\bibitem[{Jiang et~al.(2023)Jiang, Xie, Tan, Ye and Nguyen}]{jiang2023generalization}
\bibinfo{author}{Jiang, Y.}, \bibinfo{author}{Xie, J.}, \bibinfo{author}{Tan, X.}, \bibinfo{author}{Ye, N.}, \bibinfo{author}{Nguyen, Q.}, \bibinfo{year}{2023}.
\newblock \bibinfo{title}{Generalization of deep learning models for predicting spatial gene expression profiles using histology images: A breast cancer case study}.
\newblock \bibinfo{journal}{bioRxiv} \URLprefix \url{https://www.biorxiv.org/content/early/2023/09/22/2023.09.20.558624}, \DOIprefix\doi{10.1101/2023.09.20.558624}, \href{http://arxiv.org/abs/https://www.biorxiv.org/content/early/2023/09/22/2023.09.20.558624.full.pdf}{\tt arXiv:https://www.biorxiv.org/content/early/2023/09/22/2023.09.20.558624.full.pdf}.
\bibitem[{Johnson et~al.(2007)Johnson, Li and Rabinovic}]{johnson2007adjusting}
\bibinfo{author}{Johnson, W.E.}, \bibinfo{author}{Li, C.}, \bibinfo{author}{Rabinovic, A.}, \bibinfo{year}{2007}.
\newblock \bibinfo{title}{Adjusting batch effects in microarray expression data using empirical bayes methods}.
\newblock \bibinfo{journal}{Biostatistics} \bibinfo{volume}{8}, \bibinfo{pages}{118--127}.
\bibitem[{Kingma and Ba(2017)}]{kingma2017adam}
\bibinfo{author}{Kingma, D.P.}, \bibinfo{author}{Ba, J.}, \bibinfo{year}{2017}.
\newblock \bibinfo{title}{Adam: A method for stochastic optimization} \href{http://arxiv.org/abs/1412.6980}{\tt arXiv:1412.6980}.
\bibitem[{Korsunsky et~al.(2019)Korsunsky, Millard, Fan et~al.}]{korsunsky2019fast}
\bibinfo{author}{Korsunsky, I.}, \bibinfo{author}{Millard, N.}, \bibinfo{author}{Fan, J.}, et~al., \bibinfo{year}{2019}.
\newblock \bibinfo{title}{Fast, sensitive and accurate integration of single-cell data with harmony}.
\newblock \bibinfo{journal}{Nature methods} \bibinfo{volume}{16}, \bibinfo{pages}{1289--1296}.
\bibitem[{Lammi and Qu(2024)}]{lammi2024spatial}
\bibinfo{author}{Lammi, M.J.}, \bibinfo{author}{Qu, C.}, \bibinfo{year}{2024}.
\newblock \bibinfo{title}{Spatial transcriptomics, proteomics, and epigenomics as tools in tissue engineering and regenerative medicine}.
\newblock \bibinfo{journal}{Bioengineering} \bibinfo{volume}{11}, \bibinfo{pages}{1235}.
\bibitem[{Li et~al.(2024)Li, Zhang, Wang, Zhang, Li, Wang and Song}]{li2024gene}
\bibinfo{author}{Li, B.}, \bibinfo{author}{Zhang, Y.}, \bibinfo{author}{Wang, Q.}, \bibinfo{author}{Zhang, C.}, \bibinfo{author}{Li, M.}, \bibinfo{author}{Wang, G.}, \bibinfo{author}{Song, Q.}, \bibinfo{year}{2024}.
\newblock \bibinfo{title}{Gene expression prediction from histology images via hypergraph neural networks}.
\newblock \bibinfo{journal}{Briefings in Bioinformatics} \bibinfo{volume}{25}, \bibinfo{pages}{bbae500}.
\bibitem[{Lopez et~al.(2019)Lopez, Nazaret, Langevin, Samaran, Regier, Jordan and Yosef}]{lopez2019joint}
\bibinfo{author}{Lopez, R.}, \bibinfo{author}{Nazaret, A.}, \bibinfo{author}{Langevin, M.}, \bibinfo{author}{Samaran, J.}, \bibinfo{author}{Regier, J.}, \bibinfo{author}{Jordan, M.I.}, \bibinfo{author}{Yosef, N.}, \bibinfo{year}{2019}.
\newblock \bibinfo{title}{A joint model of unpaired data from scrna-seq and spatial transcriptomics for imputing missing gene expression measurements}.
\newblock \bibinfo{journal}{arXiv preprint arXiv:1905.02269} .
\bibitem[{Marel(2024)}]{marel2024navigating}
\bibinfo{author}{Marel, R.v.d.}, \bibinfo{year}{2024}.
\newblock \bibinfo{title}{Navigating the complexity of data imputation in spatial transcriptomics: Strategies, challenges, and future directions} .
\bibitem[{Mejia et~al.(2023)Mejia, C\'ardenas, Ruiz, Castillo and Arbel\'aez}]{mejia2023SEPAL}
\bibinfo{author}{Mejia, G.}, \bibinfo{author}{C\'ardenas, P.}, \bibinfo{author}{Ruiz, D.}, \bibinfo{author}{Castillo, A.}, \bibinfo{author}{Arbel\'aez, P.}, \bibinfo{year}{2023}.
\newblock \bibinfo{title}{Sepal: Spatial gene expression prediction from local graphs}, in: \bibinfo{booktitle}{Proceedings of the IEEE/CVF International Conference on Computer Vision (ICCV) Workshops}, pp. \bibinfo{pages}{2294--2303}.
\bibitem[{Mejia et~al.(2024)Mejia, Ruiz, C{\'a}rdenas, Manrique, Vega and Arbel{\'a}ez}]{mejia2024enhancing}
\bibinfo{author}{Mejia, G.}, \bibinfo{author}{Ruiz, D.}, \bibinfo{author}{C{\'a}rdenas, P.}, \bibinfo{author}{Manrique, L.}, \bibinfo{author}{Vega, D.}, \bibinfo{author}{Arbel{\'a}ez, P.}, \bibinfo{year}{2024}.
\newblock \bibinfo{title}{Enhancing gene expression prediction from histology images with spatial transcriptomics completion}, in: \bibinfo{booktitle}{International Conference on Medical Image Computing and Computer-Assisted Intervention}, \bibinfo{organization}{Springer}. pp. \bibinfo{pages}{91--101}.
\bibitem[{Mirzazadeh et~al.(2023)Mirzazadeh, Andrusivova, Larsson et~al.}]{mirzazadeh2023spatially}
\bibinfo{author}{Mirzazadeh, R.}, \bibinfo{author}{Andrusivova, Z.}, \bibinfo{author}{Larsson, L.}, et~al., \bibinfo{year}{2023}.
\newblock \bibinfo{title}{Spatially resolved transcriptomic profiling of degraded and challenging fresh frozen samples}.
\newblock \bibinfo{journal}{Nature Communications} \bibinfo{volume}{14}, \bibinfo{pages}{509}.
\bibitem[{Palla et~al.(2022)Palla, Spitzer, Klein et~al.}]{palla2022squidpy}
\bibinfo{author}{Palla, G.}, \bibinfo{author}{Spitzer, H.}, \bibinfo{author}{Klein, M.}, et~al., \bibinfo{year}{2022}.
\newblock \bibinfo{title}{Squidpy: A scalable framework for spatial omics analysis}.
\newblock \bibinfo{journal}{Nature Methods} \bibinfo{volume}{19}, \bibinfo{pages}{171–178}.
\newblock \DOIprefix\doi{10.1038/s41592-021-01358-2}.
\bibitem[{Pang et~al.(2021)Pang, Su and Li}]{pang2021leveraging}
\bibinfo{author}{Pang, M.}, \bibinfo{author}{Su, K.}, \bibinfo{author}{Li, M.}, \bibinfo{year}{2021}.
\newblock \bibinfo{title}{Leveraging information in spatial transcriptomics to predict super-resolution gene expression from histology images in tumors}.
\newblock \bibinfo{journal}{bioRxiv} , \bibinfo{pages}{2021--11}.
\bibitem[{Parigi et~al.(2022)Parigi, Larsson, Das et~al.}]{parigi2022spatial}
\bibinfo{author}{Parigi, S.M.}, \bibinfo{author}{Larsson, L.}, \bibinfo{author}{Das, S.}, et~al., \bibinfo{year}{2022}.
\newblock \bibinfo{title}{The spatial transcriptomic landscape of the healing mouse intestine following damage} \bibinfo{volume}{13}, \bibinfo{pages}{828}.
\bibitem[{Pham et~al.(2023)Pham, Tan, Balderson et~al.}]{pham2023robust}
\bibinfo{author}{Pham, D.}, \bibinfo{author}{Tan, X.}, \bibinfo{author}{Balderson, B.}, et~al., \bibinfo{year}{2023}.
\newblock \bibinfo{title}{Robust mapping of spatiotemporal trajectories and cell–cell interactions in healthy and diseased tissues}.
\newblock \bibinfo{journal}{Nature Communications} \bibinfo{volume}{14}.
\newblock \DOIprefix\doi{10.1038/s41467-023-43120-6}.
\bibitem[{Shengquan et~al.(2021)Shengquan, Boheng, Xiaoyang, Xuegong and Rui}]{shengquan2021stplus}
\bibinfo{author}{Shengquan, C.}, \bibinfo{author}{Boheng, Z.}, \bibinfo{author}{Xiaoyang, C.}, \bibinfo{author}{Xuegong, Z.}, \bibinfo{author}{Rui, J.}, \bibinfo{year}{2021}.
\newblock \bibinfo{title}{stplus: a reference-based method for the accurate enhancement of spatial transcriptomics}.
\newblock \bibinfo{journal}{Bioinformatics} \bibinfo{volume}{37}, \bibinfo{pages}{i299--i307}.
\bibitem[{Stickels et~al.(2021)Stickels, Murray, Kumar et~al.}]{slide_seq_v2}
\bibinfo{author}{Stickels, R.R.}, \bibinfo{author}{Murray, E.}, \bibinfo{author}{Kumar, P.}, et~al., \bibinfo{year}{2021}.
\newblock \bibinfo{title}{Highly sensitive spatial transcriptomics at near-cellular resolution with slide-seqv2}.
\newblock \bibinfo{journal}{Nature Biotechnology} \bibinfo{volume}{39}, \bibinfo{pages}{313--319}.
\newblock \URLprefix \url{https://www.nature.com/articles/s41587-020-0739-1}, \DOIprefix\doi{10.1038/s41587-020-0739-1}.
\bibitem[{Stuart et~al.(2019)Stuart, Butler, Hoffman et~al.}]{stuart2019comprehensive}
\bibinfo{author}{Stuart, T.}, \bibinfo{author}{Butler, A.}, \bibinfo{author}{Hoffman, P.}, et~al., \bibinfo{year}{2019}.
\newblock \bibinfo{title}{Comprehensive integration of single-cell data}.
\newblock \bibinfo{journal}{cell} \bibinfo{volume}{177}, \bibinfo{pages}{1888--1902}.
\bibitem[{Ståhl et~al.(2016)Ståhl, Salmén, Vickovic et~al.}]{10x_visium}
\bibinfo{author}{Ståhl, P.L.}, \bibinfo{author}{Salmén, F.}, \bibinfo{author}{Vickovic, S.}, et~al., \bibinfo{year}{2016}.
\newblock \bibinfo{title}{Visualization and analysis of gene expression in tissue sections by spatial transcriptomics}.
\newblock \bibinfo{journal}{Science} \bibinfo{volume}{353}, \bibinfo{pages}{78--82}.
\newblock \URLprefix \url{https://www.science.org/doi/10.1126/science.aaf2403}, \DOIprefix\doi{10.1126/science.aaf2403}.
\bibitem[{Vaswani et~al.(2023)Vaswani, Shazeer, Parmar et~al.}]{vaswani2023attention}
\bibinfo{author}{Vaswani, A.}, \bibinfo{author}{Shazeer, N.}, \bibinfo{author}{Parmar, N.}, et~al., \bibinfo{year}{2023}.
\newblock \bibinfo{title}{Attention is all you need} \href{http://arxiv.org/abs/1706.03762}{\tt arXiv:1706.03762}.
\bibitem[{Vicari et~al.(2023)Vicari, Mirzazadeh, Nilsson et~al.}]{vicari2023spatial}
\bibinfo{author}{Vicari, M.}, \bibinfo{author}{Mirzazadeh, R.}, \bibinfo{author}{Nilsson, A.}, et~al., \bibinfo{year}{2023}.
\newblock \bibinfo{title}{Spatial multimodal analysis of transcriptomes and metabolomes in tissues}.
\newblock \bibinfo{journal}{Nature Biotechnology} , \bibinfo{pages}{1--5}.
\bibitem[{Villacampa et~al.(2021)Villacampa, Larsson, Mirzazadeh et~al.}]{villacampa2021genome}
\bibinfo{author}{Villacampa, E.G.}, \bibinfo{author}{Larsson, L.}, \bibinfo{author}{Mirzazadeh, R.}, et~al., \bibinfo{year}{2021}.
\newblock \bibinfo{title}{Genome-wide spatial expression profiling in formalin-fixed tissues}.
\newblock \bibinfo{journal}{Cell Genomics} \bibinfo{volume}{1}.
\bibitem[{Wang et~al.(2023)Wang, Wu, Xiong et~al.}]{wang2023crost}
\bibinfo{author}{Wang, G.}, \bibinfo{author}{Wu, S.}, \bibinfo{author}{Xiong, Z.}, et~al., \bibinfo{year}{2023}.
\newblock \bibinfo{title}{{CROST: a comprehensive repository of spatial transcriptomics}}.
\newblock \bibinfo{journal}{Nucleic Acids Research} \bibinfo{volume}{52}, \bibinfo{pages}{D882--D890}.
\newblock \URLprefix \url{https://doi.org/10.1093/nar/gkad782}, \DOIprefix\doi{10.1093/nar/gkad782}, \href{http://arxiv.org/abs/https://academic.oup.com/nar/article-pdf/52/D1/D882/55040066/gkad782.pdf}{\tt arXiv:https://academic.oup.com/nar/article-pdf/52/D1/D882/55040066/gkad782.pdf}.
\bibitem[{Welch et~al.(2019)Welch, Kozareva, Ferreira, Vanderburg, Martin and Macosko}]{welch2019single}
\bibinfo{author}{Welch, J.D.}, \bibinfo{author}{Kozareva, V.}, \bibinfo{author}{Ferreira, A.}, \bibinfo{author}{Vanderburg, C.}, \bibinfo{author}{Martin, C.}, \bibinfo{author}{Macosko, E.Z.}, \bibinfo{year}{2019}.
\newblock \bibinfo{title}{Single-cell multi-omic integration compares and contrasts features of brain cell identity}.
\newblock \bibinfo{journal}{Cell} \bibinfo{volume}{177}, \bibinfo{pages}{1873--1887}.
\bibitem[{Xie et~al.(2023)Xie, Pang, Bader and Wang}]{xie2023spatially}
\bibinfo{author}{Xie, R.}, \bibinfo{author}{Pang, K.}, \bibinfo{author}{Bader, G.D.}, \bibinfo{author}{Wang, B.}, \bibinfo{year}{2023}.
\newblock \bibinfo{title}{Spatially resolved gene expression prediction from h\&e histology images via bi-modal contrastive learning}.
\newblock \bibinfo{journal}{arXiv preprint arXiv:2306.01859} .
\bibitem[{Yan et~al.(2024)Yan, Zhu, Chen, Yang, Cui, Zou and Zhang}]{yan2024integration}
\bibinfo{author}{Yan, C.}, \bibinfo{author}{Zhu, Y.}, \bibinfo{author}{Chen, M.}, \bibinfo{author}{Yang, K.}, \bibinfo{author}{Cui, F.}, \bibinfo{author}{Zou, Q.}, \bibinfo{author}{Zhang, Z.}, \bibinfo{year}{2024}.
\newblock \bibinfo{title}{Integration tools for scrna-seq data and spatial transcriptomics sequencing data}.
\newblock \bibinfo{journal}{Briefings in Functional Genomics} \bibinfo{volume}{23}, \bibinfo{pages}{295--302}.
\bibitem[{Yang et~al.(2024)Yang, Hossain, Stone and Rahman}]{yang2024spatial}
\bibinfo{author}{Yang, Y.}, \bibinfo{author}{Hossain, M.Z.}, \bibinfo{author}{Stone, E.}, \bibinfo{author}{Rahman, S.}, \bibinfo{year}{2024}.
\newblock \bibinfo{title}{Spatial transcriptomics analysis of gene expression prediction using exemplar guided graph neural network}.
\newblock \bibinfo{journal}{Pattern Recognition} \bibinfo{volume}{145}, \bibinfo{pages}{109966}.
\bibitem[{Yang et~al.(2023)Yang, Hossain, Stone and Rahman}]{yang2023exemplar}
\bibinfo{author}{Yang, Y.}, \bibinfo{author}{Hossain, M.Z.}, \bibinfo{author}{Stone, E.A.}, \bibinfo{author}{Rahman, S.}, \bibinfo{year}{2023}.
\newblock \bibinfo{title}{Exemplar guided deep neural network for spatial transcriptomics analysis of gene expression prediction}, in: \bibinfo{booktitle}{Proceedings of the IEEE/CVF Winter Conference on Applications of Computer Vision}, pp. \bibinfo{pages}{5039--5048}.
\bibitem[{Zeng et~al.(2022)Zeng, Wei, Yu et~al.}]{zeng2022spatial}
\bibinfo{author}{Zeng, Y.}, \bibinfo{author}{Wei, Z.}, \bibinfo{author}{Yu, W.}, et~al., \bibinfo{year}{2022}.
\newblock \bibinfo{title}{Spatial transcriptomics prediction from histology jointly through transformer and graph neural networks}.
\newblock \bibinfo{journal}{Briefings in Bioinformatics} \bibinfo{volume}{23}, \bibinfo{pages}{bbac297}.
\bibitem[{Zhang et~al.(2018)Zhang, Zhou, Lin and Sun}]{zhang2018shufflenet}
\bibinfo{author}{Zhang, X.}, \bibinfo{author}{Zhou, X.}, \bibinfo{author}{Lin, M.}, \bibinfo{author}{Sun, J.}, \bibinfo{year}{2018}.
\newblock \bibinfo{title}{Shufflenet: An extremely efficient convolutional neural network for mobile devices}, in: \bibinfo{booktitle}{Proceedings of the IEEE conference on computer vision and pattern recognition}, pp. \bibinfo{pages}{6848--6856}.

\end{thebibliography}

\end{document}